\documentclass[sigconf]{acmart}
\usepackage{natbib}
\usepackage{graphicx}
\usepackage{wrapfig}
\usepackage{pgfplots}
\pgfplotsset{compat=1.18}
\usepackage{enumitem}
\usepackage{xspace}
\usepackage{tabularx}
\usepackage{array}
\usepackage{amsmath}
\usepackage{titletoc}   
\usepackage[ruled,vlined]{algorithm2e}
\usepackage{bbding}
\usepackage{comment} 
\usepackage{makecell}
\usepackage{threeparttable}

\usepackage{multirow}
\usepackage{booktabs}
\AtBeginDocument{%
  }

\copyrightyear{2026}
\acmYear{2026}
\setcopyright{cc}
\setcctype{by}
\acmConference[KDD '26]{Proceedings of the 32nd ACM SIGKDD Conference on Knowledge Discovery and Data Mining V.2}{August 09--13, 2026}{Jeju Island, Republic of Korea}
\acmBooktitle{Proceedings of the 32nd ACM SIGKDD Conference on Knowledge Discovery and Data Mining V.2 (KDD '26), August 09--13, 2026, Jeju Island, Republic of Korea}
\acmDOI{10.1145/3770855.3817831}
\acmISBN{979-8-4007-2259-2/2026/08}

\acmSubmissionID{v2rtp1298}

\begin{document}

\title{Spectral-Inspired Neural Operator Learning with Limited Data and Unknown Physics}

\settopmatter{authorsperrow=3}

\author{Han Wan}
\email{wanhan2001@ruc.edu.cn}
\affiliation{%
  \institution{Gaoling School of Artificial Intelligence, Renmin University of China}
  \city{Beijing}
  \country{China}
}

\author{Rui Zhang}
\authornote{Corresponding author.}
\email{rayzhang@ruc.edu.cn}
\affiliation{%
  \institution{Gaoling School of Artificial Intelligence, Renmin University of China}
  \city{Beijing}
  \country{China}
}

\author{Hao Sun}
\email{haosun@ruc.edu.cn}
\affiliation{%
  \institution{Gaoling School of Artificial Intelligence, Renmin University of China}
  \city{Beijing}
  \country{China}
}

\begin{abstract}
Learning PDE dynamics from limited data with unknown physics is challenging. Existing neural PDE solvers either require large datasets or rely on known physics (e.g., PDE residuals or handcrafted stencils), leading to limited applicability. To address these challenges, we propose \underline{\textbf{S}}pectral-\underline{\textbf{I}}nspired \underline{\textbf{N}}eural \underline{\textbf{O}}perator (SINO), which can model complex systems from just 5 trajectories without requiring explicit PDE terms. Specifically, SINO automatically learns Fourier multipliers as functions of frequency indices, capturing both high-order derivatives and global/local couplings, thus enabling a compact representation of the underlying differential operators in physics-agnostic regimes. To model nonlinear effects, it employs a $\Pi$-block that performs multiplicative interactions on derivative features, followed by a spectral low-pass filter for de-aliasing. Extensive experiments on both 2D and 3D PDE benchmarks demonstrate that SINO achieves state-of-the-art performance, with improvements of 1-2 orders of magnitude in accuracy. Particularly, with only 5 training trajectories, SINO outperforms data-driven methods trained on 200 trajectories and remains predictive on challenging out-of-distribution cases where other methods fail. 
\end{abstract}

\ccsdesc[500]{Computing methodologies~Artificial intelligence}
\ccsdesc[500]{Computing methodologies~Modeling and simulation}
\ccsdesc[300]{Applied computing~Physics}

\keywords{PDE modeling, neural operators, spectral methods, AI for Physics}


\maketitle

\section{Introduction}\label{Introduction}
Partial differential equations (PDEs) govern the evolution of numerous physical systems, from heat transfer~\citep{pletcher2012computational} and chemical reactions~\citep{ghergu2011nonlinear} to fluid dynamics~\citep{holton2013introduction}. Accurately modeling and simulating such PDE-governed systems is essential for understanding natural phenomena and engineering applications. However, in many scientific computing scenarios, we face two fundamental challenges~\cite{azizzadenesheli2024neural,zhang2023artificial}: \textbf{(i) only a few trajectories of data are available}, and \textbf{(ii) the governing physics is unknown}. While traditional methods offer strong stability and interpretability, their reliance on a deep understanding of the physical system limits their application in scenarios where the underlying physics is unknown.

\sloppy In recent years, data-driven neural PDE solvers have been proposed to address the limitations of traditional numerical methods~\citep{azizzadenesheli2024neural}. Representative works include DeepONet~\citep{lu2021learning}, FNO~\citep{li2020fourier}, and DPOT~\citep{hao2024dpot}. By leveraging the expressive power of neural networks and large datasets, these methods learn mappings between function spaces. Therefore, they can reduce computational costs under coarse spatiotemporal resolutions and do not require explicit knowledge of the underlying PDEs. However, because they fit observed data with limited physical inductive biases, these methods are prone to overfitting the training distribution, resulting in high data demand and poor out-of-distribution generalization to unseen conditions. In many scientific and engineering domains, obtaining sufficient training data requires expensive physical experiments (e.g., wind tunnel testing or combustion experiments), which severely limits their practical applications~\citep{parente2024data,li2024learning,zhang2025artificial}.

\sloppy To reduce the data demands of data-driven solvers, physics-aware methods incorporate physical knowledge into neural surrogates, yielding two main paradigms: physics-informed and physics-encoded~\citep{faroughi2022physics}. Physics‑informed methods, such as PINNs~\citep{raissi2019physics} and PINO~\citep{li2024physics}, add soft constraints by including PDE residuals in the loss function. They require explicit forms of the governing PDEs and often suffer from training instability~\citep{krishnapriyan2021characterizing}. On the other hand, physics-encoded methods, like PeRCNN~\citep{Rao_2023, rao2022discovering} and TSM~\citep{sun2023neural}, embed physical rules into the network architecture, e.g., hardcoding finite difference stencils as convolutional kernels~\citep{Rao_2023}. When designing these models, one still needs complete or partial PDE terms as prior knowledge. Moreover, since many architectures borrow from local numerical schemes such as the finite difference method (FDM) or the finite volume method (FVM), they inherit limitations in handling high-order derivatives and global interactions~\citep{mcgreivy2024weak}.

\begin{figure}
    \centering
    \includegraphics[width=0.95\linewidth]{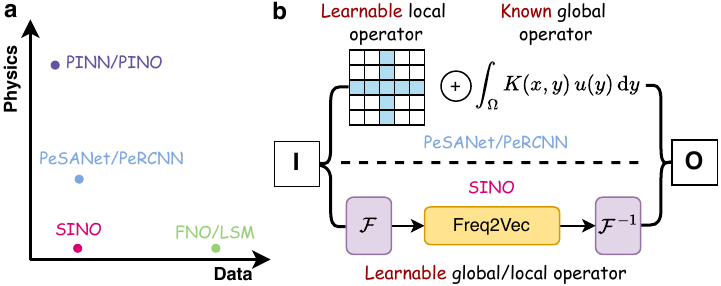}
\caption{Motivation of SINO. a. SINO targets operator learning in the limited-data, unknown-physics regime. b. Unlike physics-aware methods that combine a learnable local term with a known global operator, SINO learns the global/local interactions in Fourier domain via Freq2Vec.}
    \label{fig:shiyi}
    \vspace{-8pt}
\end{figure}

To address these challenges, we propose the \underline{\textbf{S}}pectral-\underline{\textbf{I}}nspired \underline{\textbf{N}}eural \underline{\textbf{O}}perator (SINO), which can accurately model complex systems from as few as 5 trajectories without any explicit PDE terms. The design of SINO is inspired by spectral methods, which are well known for their ability to capture global information with exponential accuracy~\citep{JML-1-3}. Specifically, the Frequency-to-Vector (Freq2Vec) module maps each frequency index into a learnable embedding, mimicking the derivative multipliers in spectral methods without relying on predefined operators. To further capture nonlinear dynamics, we employ a $\Pi$-block that models multiplicative interactions, together with a low-pass filter to mitigate aliasing effects. By inheriting the structure of spectral methods, SINO achieves stronger generalization from limited observations and demonstrates robust out-of-distribution (OOD) performance compared to data-driven approaches, while also handling global interactions and high-order derivatives more effectively than other physics-aware methods. Fig.~\ref{fig:shiyi} provides a positioning overview of SINO with data-driven and physics-aware methods.

\textbf{In summary, we make the following contributions:}

\textbf{(1) Novel architecture}. We propose SINO, a data-efficient neural operator for modeling complex spatiotemporal dynamics without requiring prior knowledge of PDE terms. We propose two key modules: Freq2Vec simulates Fourier multipliers as functions of frequency indices, and the nonlinear operator block captures complex interactions through efficient multiplicative operations.

\textbf{(2) State-of-the-art (SOTA) performance}. SINO achieves SOTA performance across multiple 2D and 3D PDE benchmarks, delivering improvements of 1–2 orders of magnitude in accuracy compared to baselines. To our knowledge, SINO is among the first physics-aware methods capable of simulating globally coupled systems such as the NSEs without requiring explicit PDE terms.

\textbf{(3) Data efficiency and OOD handling}. Due to the spectral-inspired design, SINO captures the underlying operator beyond memorizing training trajectories. With only 5 training trajectories, SINO outperforms data-driven methods trained on 200 trajectories and remains predictive in out-of-distribution (OOD) tests where other methods fail.

\section{Related Work}
\subsection{Data-driven Neural PDE Solvers}
When provided with abundant data, data-driven methods learn mappings between function spaces, bypassing the need for explicit PDE formulation. Representative examples include the FNO~\citep{li2020fourier} and its variants~\citep{tran2023factorized, wen2022u}, DeepONet~\citep{lu2021learning} and its variants~\citep{venturi2023svd,lee2023hyperdeeponet}. Transformer-based models~\citep{wu2024transolver,li2024scalable} and graph-based architectures~\citep{pfaff2020learning,brandstetter2022message} extend neural PDE solvers to handle complex geometries by operating on mesh or graph representations. Recently, PDE foundation models aim to solve multiple PDE families within a single framework, achieving broad applicability across different physical systems~\citep{hao2024dpot,zhang2024deciphering,hang2024unisolver}. Thanks to their rapid inference ability, these data-driven solvers have found broad scientific computing applications, such as weather forecasting~\citep{zhang2023skilful}, control systems~\citep{hu2025wavelet}, and geometric design~\citep{wang2024beno,sun2026geometry}. However, a major drawback of data-driven PDE solvers is their heavy reliance on large datasets, which in many applications are costly to obtain through expensive physical experiments~\citep{parente2024data,li2024learning}. In data-scarce regimes, they tend to overfit the training distribution, leading to degraded generalization and often poor OOD performance under shifts from the training distributions. In contrast, SINO uses a spectral-inspired framework to embed physical structure, enabling effective training from limited data and robust OOD generalization.

\subsection{Physics-aware Neural PDE Solvers}
Unlike data-driven methods, physics-aware methods aim to reduce data dependency by embedding physical priors into the learning process or architecture design~\citep{faroughi2022physics}. Based on how they incorporate physical laws, physics-aware methods can be categorized into physics-informed and physics-encoded methods. Physics-informed methods integrate PDE residuals~\citep{wang2021learning} or numerical schemes (e.g., FDM~\citep{huang2023neuralstagger}, spectral methods~\citep{du2024neural}, particle methods~\citep{zhang2025monte}) into the loss function, but require complete knowledge of the governing PDEs and often suffer from training instability~\citep{krishnapriyan2021characterizing}. Physics-encoded methods incorporate physical structure in a hard-constrained manner, such as using FDM~\citep{Long_2019,Rao_2023} or FVM~\citep{kochkov2021machine,sun2023neural}, yet depend on specific PDE terms and are constrained by their local receptive field. Additionally, recent physics-encoded approaches that incorporate spectral methods have shown promise in improving global modeling but still require full or partial knowledge of the governing PDE terms, especially when tackling complex systems such as the NSEs~\citep{dresdner2023learning, li2025symbolic}. Unlike these hybrid approaches, SINO not only captures global information in the frequency domain but also requires no prior knowledge of specific PDE terms. 

\section{Methodology}\label{Method}
\begin{figure*}
    \centering
    \includegraphics[width=0.92\linewidth]{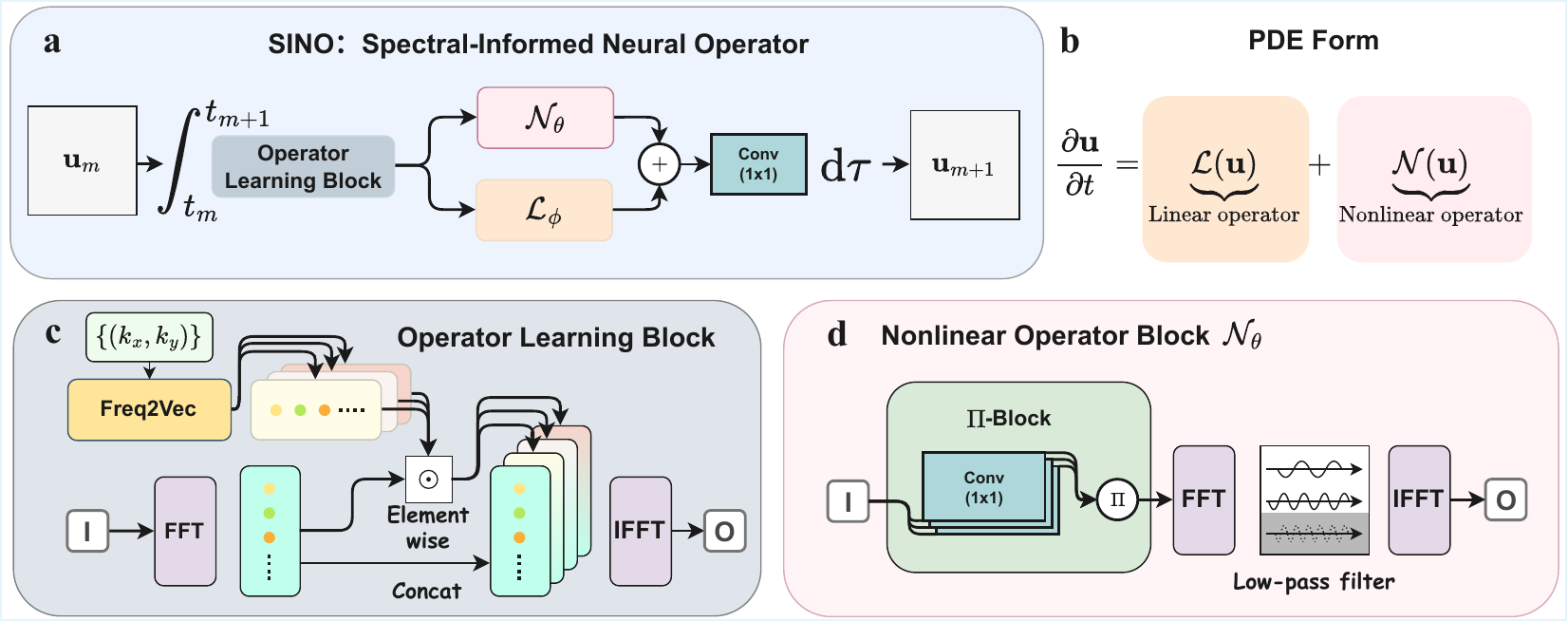}
\vspace{-3pt}
\caption{\textbf{The framework of SINO.} (a) The overall architecture. (b) The PDE formulation. (c) The structure of the Spectral Learning Block (SLB). (d) The structure of the Nonlinear Operator Block ($\Pi$-Block + low-pass filter).}
\label{fig:main}
\vspace{-2pt}
\end{figure*}

\subsection{Problem Setting and Preliminary}
This paper addresses the problem of modeling PDE-governed systems from limited data when PDE terms are unknown. Consider a system defined on $\Omega \subset \mathbb{R}^d$, whose state $\mathbf{u}(\mathbf{x},t;\boldsymbol{\mu})\in\mathbb{R}^C$ satisfies
\begin{equation}
    \partial_t \mathbf{u}(\mathbf{x},t;\boldsymbol{\mu}) = \mathcal{L}(\mathbf{u}(\mathbf{x},t;\boldsymbol{\mu})) + \mathcal{N}(\mathbf{u}(\mathbf{x},t;\boldsymbol{\mu})).
    \label{eq:pde}
\end{equation}

Here, $\mathbf{u}$ denotes the state field, $\boldsymbol{\mu}$ represents PDE parameters, and $\mathcal{L}(\cdot)$ and $\mathcal{N}(\cdot)$ are \textit{unknown} linear and nonlinear differential operators, which include multiple spatial derivatives of various orders. 
When the PDE formulation in Eq.~\ref{eq:pde} is known, spectral methods are among the most accurate and fastest numerical methods on regular domains because they exhibit exponential convergence and capture global information~\citep{shen2011spectral,mcgreivy2024weak}. Given the state $\mathbf{u}_m(\mathbf{x})$ at time $t_{m}$, a spectral solver advances to $t_{m+1}$ in three steps. 

{\textit{Firstly,}} the state is transformed to frequency space, and all spatial derivatives are computed. In detail, the Fourier transform $\mathcal{F}$ converts $\mathbf{u}_m$ into spectral coefficients $\widehat{\mathbf{u}_m}$ as follows:
\begin{equation}
        \widehat{\mathbf{u}_m}(\mathbf{k})
     = \mathcal{F}\bigl[\mathbf{u}_m(\mathbf{x})\bigr]
     = \int_{\Omega}\mathbf{u}_m(\mathbf{x})\,e^{-\,\mathrm{i}\,\mathbf{k}\cdot\mathbf{x}}\,\mathrm{d}\mathbf{x},
\end{equation}
where $\mathbf{k}$ is the frequency index corresponding to different frequency components. In the frequency domain, each derivative operator can become an exact multiplier. For example, the Laplacian operator $\nabla^2 \mathbf{u}_m$ satisfies $\widehat{\nabla^2 \mathbf{u}_m}(\mathbf{k}) = -\|\mathbf{k}\|_{2}^{2}\,\widehat{\mathbf{u}_m}(\mathbf{k})$ in the frequency domain. This suggests learning multipliers as functions of $\mathbf{k}$ is a compact way to represent unknown differential operators.

{\textit{Secondly,}} each derivative is transformed back to the spatial domain via the inverse Fourier transform $\mathcal{F}^{-1}$ and combined to form the linear and nonlinear terms of the PDE. For linear terms, the corresponding spectral derivatives can be directly scaled by their coefficients and then summed. For example, the diffusion term $\nu \nabla^2 \mathbf{u}_m$ can be denoted as $\nu\mathcal{F}^{-1}[\widehat{\nabla^2 \mathbf{u}_m}]$. For nonlinear terms, the computation requires calculating element-wise products in the spatial domain. For instance, the convection term $\nabla \mathbf{u}_m\cdot \mathbf{u}_m$ can be calculated as $\mathcal{F}^{-1}[\widehat{\nabla \mathbf{u}_m}] \cdot \mathcal{F}^{-1}[\widehat{\mathbf{u}_m}]$.

{\textit{Thirdly,}} a high‑order time‑stepping scheme (e.g., a fourth-order Runge-Kutta (RK4) scheme) is deployed to obtain $\mathbf{u}_{m+1}(\mathbf{x})$.

\subsection{Overall Architecture}
Motivated by spectral methods, we propose SINO to model nonlinear systems from limited data when all PDE terms are unknown. Similar to a spectral solver, SINO follows three steps (Fig.~\ref{fig:main}a). 

\textit{Firstly,} SINO maps the input state $\mathbf{u}_m$ to the frequency domain via an appropriate spectral transform (e.g., Fourier/cosine/sine), chosen according to the underlying boundary conditions, and uses a Spectral Learning Block ($\text{SLB}$) to simulate various spatial derivatives (Fig.~\ref{fig:main}c). Because the exact PDE terms are unknown, we employ a Frequency-to-Vector (Freq2Vec) module that learns how to represent each derivative from its frequency index $\mathbf{k}$ under the supervision of limited data.

\textit{Secondly,} the learned representations are transformed through linear ($\mathcal{L}_{\phi}$) and nonlinear ($\mathcal{N}_{\theta}$) parts and then recombined by a $1\times 1$ convolution, where $\phi$ and $\theta$ denote the parameters of SINO. Specifically, in the nonlinear branch, we use a $\Pi$-block to model the nonlinear interactions via element-wise products and a low‑pass filter to prevent aliasing (Fig.~\ref{fig:main}d). 

\textit{Thirdly,} the combined right‑hand side (RHS) is advanced in time using an RK4 scheme to obtain $\mathbf{u}_{m+1}$. As a result, the time advancing of SINO can be described as

\begin{equation}
\begin{aligned}
\mathbf{h} &= \mathrm{SLB}(\mathbf{u}), \\
\mathrm{RHS}(\mathbf{u}) &= \mathrm{Conv}_{1\times 1}\!\left(
\mathcal{L}_{\phi}(\mathbf{h})+\mathcal{N}_{\theta}(\mathbf{h})
\right), \\
\mathbf{u}_{m+1} &= \mathrm{RK4}\!\left(\mathbf{u}_m, \mathrm{RHS}, \Delta t\right),
\qquad \Delta t = t_{m+1}-t_m .
\end{aligned}
\label{eq:SINO}
\end{equation}

\subsection{Architecture Components of SINO}
We herein detail the network architecture of SINO, including the spectral learning block, the linear block, and the nonlinear block.

\textbf{Spectral learning block.} In spectral methods, spatial derivatives are computed exactly in the frequency domain, such as $\widehat{\partial_{x}^n \mathbf{u}}(\mathbf{k}) = (\mathrm{i}{k}_x)^n \cdot \widehat{\mathbf{u}}({k}) 
$. When PDE terms are unknown, we design a spectral learning block to learn various spatial derivatives in the frequency domain. In detail, we propose a key module named Frequency-to-Vector (Freq2Vec). This module learns a latent representation $\psi(\mathbf{k})$ for each frequency index $\mathbf{k}$ via a shared MLP, i.e., $\psi(\mathbf{k}) = \mathrm{MLP}(\mathbf{k}).$
The output $\psi(\mathbf{k})$ encodes the operator action at index $\mathbf{k}$ and plays a role analogous to the exact multipliers (e.g., $(\mathrm{i}{k}_x)^n$) used in classical spectral methods. To approximate the unknown derivative operator $\mathcal{D}$, SINO computes an element-wise product in the frequency domain, i.e., 
\begin{equation}
    \widehat{\mathcal{D}\mathbf{u}}(\mathbf{k}) = \psi(\mathbf{k}) \cdot \widehat{\mathbf{u}}(\mathbf{k}), 
\end{equation}
which mimics the way spectral methods apply multipliers to represent differential operators. The result is then transformed back to the spatial domain: 
\begin{equation}
    \mathcal{D} \mathbf{u}(\mathbf{x}) = \mathcal{F}^{-1}[\widehat{\mathcal{D}\mathbf{u}}(\mathbf{k})]. 
\end{equation}

A key advantage of this spectral formulation is that SINO not only learns standard spatial differential operators with high accuracy, but also naturally captures globally coupled relations. For instance, the velocity-vorticity formulation in 2D NSE can be expressed in a simple closed form. Given vorticity $\omega = \partial_x \mathbf{u}_y - \partial_y \mathbf{u}_x$, the velocity can be recovered via the Biot--Savart law~\citep{saffman1995vortex}:
\begin{equation}
    \widehat{\mathbf{u}}(\mathbf{k}) = \frac{\mathrm{i}\,\mathbf{k}^\perp}{\|\mathbf{k}\|_2^2}\,\widehat{\omega}(\mathbf{k}), 
\quad \mathbf{k}^\perp := (-{k}_y,\,{k}_x),\label{eq:bslaw}
\end{equation}
which corresponds to a simple spectral multiplier in the frequency domain. As a result, such relations can in principle be approximated by Freq2Vec due to its expressive capacity. By contrast, other physics-encoded methods~\citep{WangEtAl2024NeurIPS, yan2025learnable} built on local discretization schemes must solve a Poisson equation in Eq.~\ref{eq:bslaw}, thereby requiring detailed PDE terms.

\textbf{Linear operator block.} After computing the spectral derivatives via the spectral learning block, we use a linear operator $\mathcal{L}_{\phi}$ to combine them with simple linear channel mixing. Concretely, the set of derived features is passed through a $1\times 1$ convolution that learns to weight and sum these channels.

\textbf{Nonlinear operator block.} Unlike the linear block’s simple channel mixing, the nonlinear operator block uses multiplicative interactions to emulate nonlinear PDE terms (Fig.~\ref{fig:main}d). Given the calculated spatial derivative features $\mathcal{D}\mathbf{u}(\mathbf{x})$, we design the $\Pi$‑block as follows. We first map the derivative features into $P$ channels via a $1\times 1$ convolution and then perform element-wise products across these channels to model nonlinear interactions:
\begin{equation}
    \mathbf{v}_{\Pi}(\mathbf{x}) = \prod_{p=1}^P \left(W_{p}\, \mathcal{D}\mathbf{u}(\mathbf{x}) + b_{p}\right),
\end{equation}
where $W_p$ and $b_p$ denote the weights and biases. This $\Pi$‑block offers a lightweight yet powerful mechanism for modeling nonlinear interactions through element‑wise products, which can naturally express various nonlinear terms (e.g., the convection term $\mathbf{u}\cdot\nabla\mathbf{u}$). As shown later in Fig.~\ref{fig:operator_features}, feature maps from the $\Pi$-block closely align with ground-truth differential operators of the underlying PDEs, suggesting that SINO implicitly learns physically meaningful operators.

However, nonlinear interactions can create frequencies above the Nyquist limit, causing aliasing, where high frequencies fold into low ones and introduce spurious noise~\citep{kravchenko1997effect}. To mitigate this effect, we perform de-aliasing in the frequency domain using the classical low-pass filter with a $2/3$-rule. Specifically, we apply a Fourier transform to the nonlinear output and zero out high-frequency modes exceeding the $2/3$ cutoff:
\begin{equation}
    \widetilde{\mathbf{v}}_{\Pi}(\mathbf{k}) = \mathbf{M}(\mathbf{k}) \cdot \mathcal{F}[\mathbf{v}_{\Pi}(\mathbf{x})](\mathbf{k}), \ \mathbf{M}(\mathbf{k}) = \begin{cases} 1, & \|\mathbf{k}\|_\infty \le 2/3 {k}_{\max}, \\ 0, & \text{otherwise}, \end{cases}
\end{equation} 
where ${k}_{\max}$ denotes the highest frequency mode. This ensures that the nonlinear output does not distort the solution with aliased frequencies. Finally, we map the de-aliased representation back to the spatial domain. By combining $\Pi$-block with the low-pass filter, the nonlinear block in SINO captures complex operator interactions efficiently and enjoys strong numerical stability.

\subsection{Relation to Other Neural Spectral Methods}

SINO differs from existing spectral methods in both \emph{what} it learns in the frequency domain and \emph{how} it introduces nonlinearity. In the frequency domain, FNO~\citep{li2020fourier} uses a large spectral kernel to mix coefficients directly; SNO~\citep{fanaskov2023spectral}, AFNO~\citep{guibas2021adaptive}, and DPOT~\citep{hao2024dpot} apply MLPs directly to frequency-domain feature maps, transforming the learned coefficients with a black-box mapping; LSM~\citep{wu2023LSM} projects the dynamics into a compact latent space and performs its basis operations. Furthermore, these designs introduce nonlinearity via generic activation functions, which improves expressiveness but is not explicitly aligned with the algebraic form of nonlinear PDE terms, thereby providing limited physical interpretability. As a result, they largely behave as black boxes in the spectral domain and can struggle in few-shot settings. In contrast, SINO learns embeddings over frequency indices $\mathbf{k}$ with Freq2Vec, encoding the prior that derivative multipliers are functions of $\mathbf{k}$. Notably, SINO primarily introduces nonlinearity through a simple multiplicative $\Pi$-block, which mirrors the structural form of common nonlinear PDE operators (e.g., advection terms $\mathbf{u}\cdot \nabla \mathbf{u}$). Consequently, SINO's intermediate features align well with the ground-truth physical operators such as $\partial_x \omega$ and $\Delta \omega$ (Fig.~\ref{fig:operator_features}), thereby yielding strong inductive biases for learning the underlying physics.

\section{Theoretical Insights of SINO}
\label{sec:theory}

In this section, we provide theoretical insights into why SINO performs well with limited data, focusing on its approximation ability and generalization behavior. In essence, SINO is a \textbf{compact yet expressive} neural operator that balances approximation accuracy and generalization. For a physical field $\mathbf{u}$, one step of SINO corresponds to predicting the right-hand side (RHS) of the PDE, which involves linear and nonlinear combinations of spatial derivatives of $\mathbf{u}$. The following theorem reveals that SINO can approximate a broad class of such operators (proof in Appendix~\ref{app:proof}):

\begin{theorem}\label{thm}
Let $\Omega=[0,1)^d$ be the $d$-torus and define
\[
\mathcal U=\Bigl\{\mathbf u:\Omega\to\mathbb R^C \ \Big|\ 
\mathbf u\ \text{is periodic},\ \|\mathbf u\|_{L^2}\le B,\ 
\|\mathbf k\|_\infty\le k_{\max}\Bigr\}.
\]
For any bounded Fourier multipliers $\psi_{j,i}:\mathcal K\to\mathbb C$ on
$\mathcal K:=\{\mathbf{k}\in\mathbb Z^d:\|\mathbf{k}\|_\infty\le k_{\max}\}$,
define the Fourier-multiplier operators
\[
\widehat{T_{\psi_{j,i}}\mathbf{u}}(\mathbf{k}) = \psi_{j,i}(\mathbf{k})\,\widehat{\mathbf{u}}(\mathbf{k}),\qquad \mathbf{k}\in\mathcal K.
\]
Consider the RHS operator
\[
\mathcal R(\mathbf{u}) \ :=\ \sum_{j=1}^J \ \prod_{i=1}^{n_j} T_{\psi_{j,i}}\mathbf{u},
\]
where the products are taken pointwise in $\mathbf{x}$. Then for any $\epsilon>0$, there exists a SINO model $\mathcal S_\theta$ such that
\[
\sup_{\mathbf{u}\in\mathcal U}\ \big\|\mathcal S_\theta(\mathbf{u})-\mathcal R(\mathbf{u})\big\|_{L^2} \ <\ \epsilon.
\]
\end{theorem}

Theorem~\ref{thm} shows that SINO can approximate a broad class of nonlinear PDE operators constructed from linear derivatives and finite multiplicative interactions, which covers various reaction-advection-diffusion systems. Beyond approximation, SINO generalizes well in low-data regimes due to its structured spectral parameterization and stability-inducing design choices. In our case, Freq2Vec parameterizes Fourier multipliers through a \emph{shared} network, which couples parameters across frequency modes and reduces the degrees of freedom compared to learning independent complex weights per mode. This acts as an implicit regularizer in the few-trajectory regime. Moreover, the $\Pi$-block explicitly captures multiplicative nonlinear interactions, which is more physically consistent than introducing nonlinearity solely through generic activation functions, improving long-horizon stability and thereby generalization to unseen conditions.

\section{Experiments}\label{Experiments}

Given 5 observed trajectories, we evaluate SINO against baselines across various 2D and 3D scenarios without any known PDE terms. We also assess its OOD generalization and conduct ablation studies. \textbf{Source code will be released upon acceptance.}

\subsection{Datasets and Baseline Models}
\textbf{Datasets.} We evaluate SINO on three PDE classes, each presenting distinct computational challenges. The detailed experimental configurations, including domain specifications, trajectory counts, and resolutions, are summarized in Table~\ref{tab:dataset_summary}.

\begin{table}[!t]
\centering
\caption{\textbf{Experimental settings.} “Gen.”: high-fidelity spectral solver; “Train.”: operator-learning resolution/time step.
For NSE, train on 10\,s and test rollouts to 15\,s.}
\vspace{-3pt}
\setlength{\tabcolsep}{2.5pt}
\renewcommand{\arraystretch}{1.15}
\begin{tabular}{lcccc}
\toprule
\textbf{Setting} & \textbf{KSE} & \textbf{NSE} & \textbf{2D Burgers} & \textbf{3D Burgers} \\
\midrule
$\Omega$ &
\shortstack{$[0,12\pi)^2$} &
\shortstack{$[0,1)^2$} &
\shortstack{$[0,2\pi)^2$} &
\shortstack{$[0,2\pi)^3$} \\
$T$ &
5 &
\shortstack{10 / 15 } &
2 &
5 \\
\# traj. (tr/va/te) &
5/2/5 & 5/2/5 & 5/2/5 & 5/2/5 \\
\midrule
Gen. $N$ &
$108^2$ & $256^2$ & $512^2$ & $128^3$ \\
Gen. $\Delta t$ &
$10^{-4}$ & $10^{-4}$ & $10^{-3}$ & $5{\times}10^{-3}$ \\
Train. $N$ &
$54^2$ & $64^2$ & $64^2$ & $64^3$ \\
Train. $\Delta t$ &
$5{\times}10^{-2}$ & $5{\times}10^{-2}$ & $5{\times}10^{-2}$ & $5{\times}10^{-2}$ \\
\bottomrule
\end{tabular}
\vspace{-2pt}
\label{tab:dataset_summary}
\end{table}

\begin{table*}[t!]
\centering
\caption{\textbf{Relative $\ell_2$ error on 8 cases.}
Best results are \textbf{bold} and second-best are \underline{underlined}.
NaN: not a number; NA: not applicable.}
\vspace{-2pt}
\setlength{\tabcolsep}{4pt}
\renewcommand{\arraystretch}{1.18}
\begin{tabular*}{0.76\textwidth}{@{\extracolsep{\fill}} l c cccc ccc}
\toprule
\multirow{2}{*}[-0.3ex]{\textbf{Model}}
& \textbf{KSE} 
& \multicolumn{4}{c}{\textbf{NSE}} 
& \multicolumn{3}{c}{\textbf{Burgers}} \\
\cmidrule(lr){2-2}\cmidrule(lr){3-6}\cmidrule(lr){7-9}
& \textbf{E1}
& \makecell{\textbf{E2}\\{$10^{-4},\,f_1$}}
& \makecell{\textbf{E3}\\{$10^{-5},\,f_1$}}
& \makecell{\textbf{E4}\\{$10^{-4},\,f_2$}}
& \makecell{\textbf{E5}\\{$10^{-5},\,f_2$}}
& \makecell{\textbf{E6}\\{2D}}
& \makecell{\textbf{E7}\\{3D}}
& \makecell{\textbf{E8}\\{Mixed BC}} \\
\midrule

\multicolumn{9}{l}{\textbf{\textit{Physics-aware methods}}}\\
P$^2$C$^2$Net~\citep{WangEtAl2024NeurIPS}   
& 0.2754 & 1.3139 & 1.1842 & 0.9675 & 0.9801 & \underline{0.0859} & NA & 0.1955 \\
PeRCNN~\citep{Rao_2023}
& 0.3205 & 0.8175 & 0.8356 & 0.9153 & 0.8356 & 0.1793 & 0.9530 & 0.2272 \\
PeSANet~\citep{wan2025pesanet}
& 0.2028 & NaN & NaN & NaN & 0.9705 & 0.2667 & NaN & 0.3398 \\

\midrule
\addlinespace[2pt]
\multicolumn{9}{l}{\textbf{\textit{Data-driven methods}}}\\
FNO~\citep{li2020fourier}
& 0.0780
& \underline{0.0558} & \underline{0.1767} & \underline{0.1156} & \underline{0.1875}
& 0.3616 & 1.0073 & 0.4113 \\
FFNO~\citep{tran2023factorized}
& 0.1081 & 0.8081 & 0.8872 & 1.0592 & 1.1909 & 0.2351 & 0.5949 & 0.2988 \\
FactFormer~\citep{li2024scalable}
& 0.2438 & 0.5714 & 0.7139 & 0.2597 & 0.2134 & 0.1993 & 0.9459 & 0.2069 \\
CNext~\citep{ohana2024well}
& 0.1797 & 0.7441 & 0.8606 & 0.3012 & 0.3288 & 0.2205 & \underline{0.3654} & \underline{0.1803} \\
LSM~\citep{wu2023LSM}
& 0.1758 & 0.7254 & 0.8262 & 0.3558 & 0.3758 & 0.2104 & NA & 0.2085 \\
SNO~\citep{fanaskov2023spectral}
& 0.7985 & 1.0688 & 1.0400 & 0.7505 & 1.1140 & 0.5536 & NA & 0.5883 \\
NIPs~\citep{pmlr-v267-liu25cp}
& \underline{0.0296} & 0.2974 & 0.6123 & 0.5490 & 5.3046 & 0.3906 & NA & 0.5480 \\
RNO~\citep{liu2026riesz}
& 0.1062 & 0.1628 & 0.5145 & 0.3600 & 0.3460 & 0.3656 & NA & 0.4186 \\

\midrule
\textbf{SINO (ours)}
& \textbf{0.0040}
& \textbf{0.0171} & \textbf{0.0199} & \textbf{0.0049} & \textbf{0.0132}
& \textbf{0.0110} & \textbf{0.0097} & \textbf{0.0376} \\
\bottomrule
\end{tabular*}
\vspace{-4pt}
\label{tab:model_performance}
\end{table*}

\begin{figure*}[!t]
\centering
\includegraphics[width=0.92\linewidth]{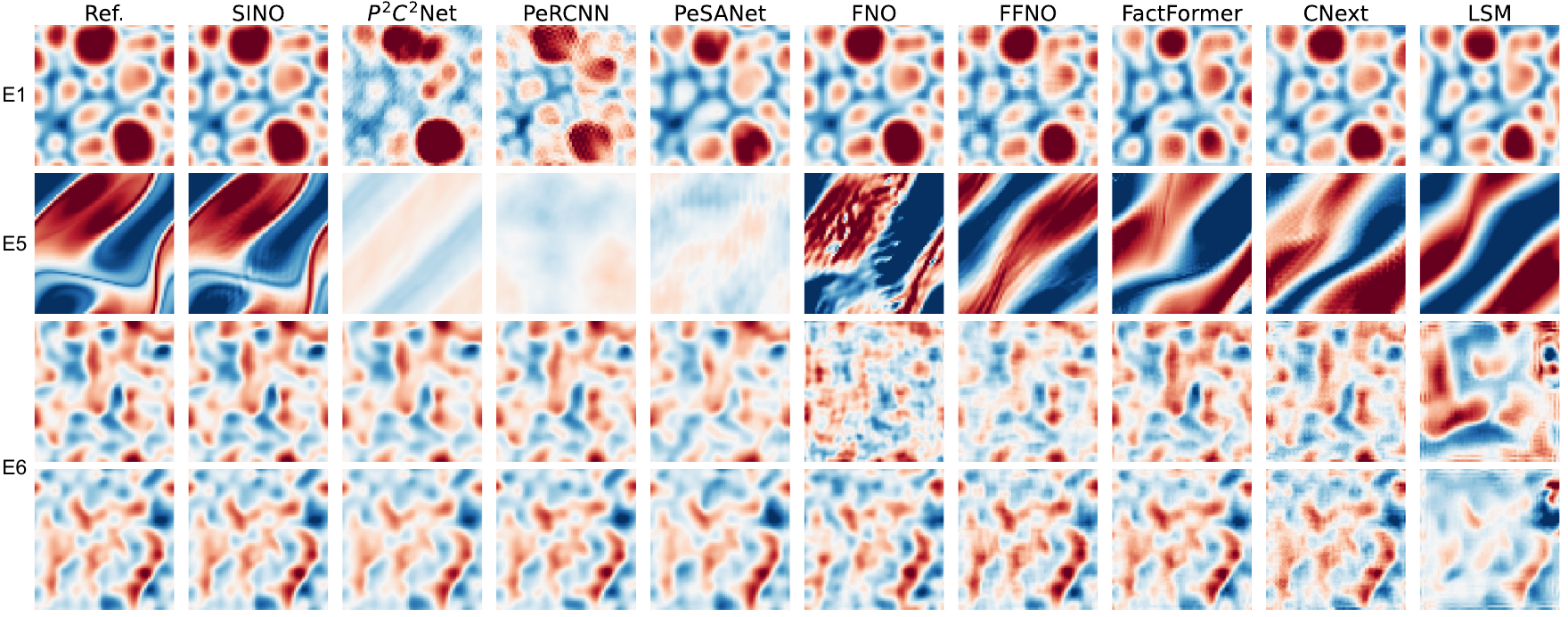}
\caption{\textbf{Predicted snapshots of SINO and selected baselines on representative 2D cases.} 
We select E1 (KSE), E5 (NSE), and E6 (2D Burgers, $\mathbf{u} = [u,v]$) as representative examples to visualize the final prediction results.}
\label{fig:mainresult}
\vspace{-1pt}
\end{figure*}

\begin{figure*}[!t]
\centering
\includegraphics[width=0.9\linewidth]{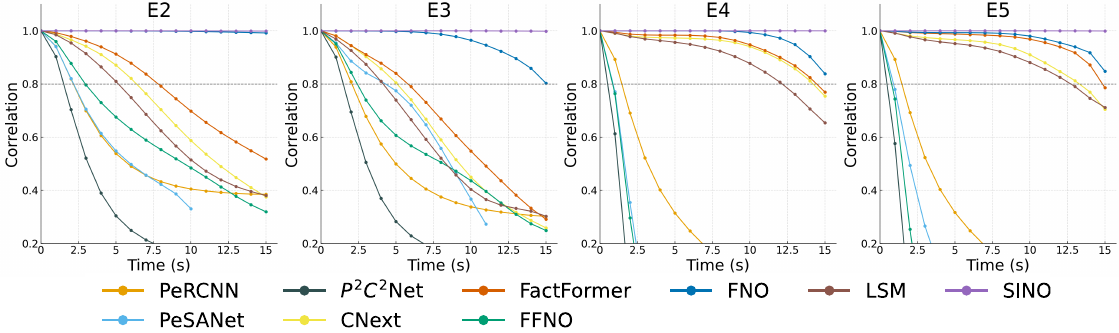}
\caption{{Error evaluation of correlation across NSE cases E2--E5.} Training uses the first 10~s; testing rolls out to 15~s.}
\label{fig:corr}
\end{figure*}

\begin{figure}[!t]
\centering
\vspace{-1pt}
\includegraphics[width=1\linewidth]{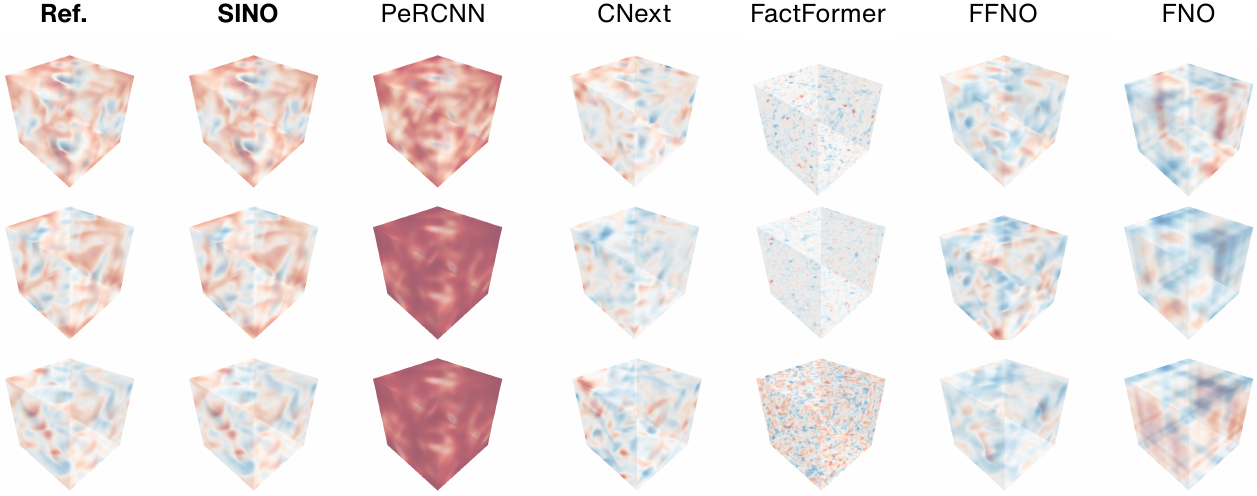}
\caption{\textbf{Predicted snapshots of SINO and other baselines on 3D Burgers cases (E7).} The three rows denote the three components of $\mathbf{u} = [u, v, w]$, respectively.}
\label{fig:3dmainresult}
\vspace{-1pt}
\end{figure}

\textbf{(1) Kuramoto-Sivashinsky Equation (KSE).} We consider the 2D KSE~\citep{jayaprakash1993universal}:
\begin{equation}
\partial_t {u} = -\nabla^2 {u} - \nabla^4 {u} - 0.5 |\nabla {u}|^2,\quad \mathbf{x}\in[0,12\pi)^2,\ t\in [0,5].
\end{equation}
The KSE is known for its stiffness due to a biharmonic term $\nabla^4{u}$ and the emergence of spatiotemporal chaos. We use this PDE to evaluate the performance of SINO in handling high-order derivatives and denote the corresponding experiment as E1.

\textbf{(2) Navier-Stokes Equation (NSE).} To examine globally coupled nonlinear dynamics and assess extrapolation ability, we adopt the 2D incompressible NSE in vorticity form as follows:
\begin{equation}\label{eq:nse}
{\partial_t \omega} = \nu \nabla^2 \omega - (\mathbf{u} \cdot \nabla)\omega + {f},\quad \mathbf{x}\in[0,1)^2,\ t\in [0,15],
\end{equation}
where $\omega$ is the vorticity and $\mathbf{u}$ is the velocity. NSE presents several challenges like multi-scale turbulence and nonlocal coupling between $\omega$ and $\mathbf{u}$. We consider two viscosities ($\nu \in \{10^{-4}, 10^{-5}\}$) and two forcings ($f_1(\mathbf{x}) = 0.1\cos(8\pi{x}_1), \ f_2(\mathbf{x}) = 0.1\sqrt{2}\,\sin\!\left(2\pi({x}_1+{x}_2) + \tfrac{\pi}{4}\right)$), yielding four setups (E2-E5). To evaluate extrapolation ability, we provide the first 10~s of data during training, while testing extends to 15~s.

\textbf{(3) Burgers' equation.} 
We consider Burgers' equation, a canonical model for nonlinear advection and shock formation:
\begin{equation}
\partial_t \mathbf{u} = 0.01 \nabla^2 \mathbf{u} - (\mathbf{u} \cdot \nabla)\mathbf{u}, 
\quad \mathbf{x}\in[0,2\pi)^d,\ t\in [0,2],
\end{equation}
where $d$ denotes the spatial dimension. In our experiments, we consider the fully periodic 2D ($d=2$, E6) and 3D ($d=3$, E7) cases. {Additionally, we introduce a 2D case with mixed boundary conditions (E8): periodic in the $x$-direction and homogeneous Dirichlet in the $y$-direction ($\mathbf{u}|_{y=0,2\pi}=\mathbf{0}$).}

All datasets are simulated with high-resolution spectral solvers. We subsample 5 low-resolution trajectories per case for training. 
\textbf{Baseline models.} We compare SINO with data-driven baselines including \textbf{FNO}~\citep{li2020fourier}, \textbf{FFNO}~\citep{tran2023factorized}, \textbf{FactFormer}~\citep{li2024scalable}, \textbf{CNext}~\citep{liu2022convnet,ohana2024well}, \textbf{LSM}~\citep{wu2023LSM}, \textbf{SNO}~\citep{fanaskov2023spectral}, 
\textbf{NIPs}~\citep{pmlr-v267-liu25cp}, and \textbf{RNO}~\citep{liu2026riesz}. 
We also include physics-encoded models such as \textbf{P$^2$C$^2$Net}~\citep{WangEtAl2024NeurIPS} , \textbf{PeRCNN}~\citep{Rao_2023} and \textbf{PeSANet}~\citep{wan2025pesanet}. We use the same train/val/test splits, input/output resolution, rollout horizon, and coarse-grid setting (large $\Delta x$ and effective time step) for all methods.
We adopt official implementations and recommended training setups when available, and select hyperparameters via validation-based grid search (Appendix~\ref{Baseline Models}). All models are trained under a unified protocol (optimizer/schedule, training iterations). Runtime is measured on the same GPU with an identical batch size and rollout length (Appendix~\ref{Training Details}).

\subsection{Main Results}

In this section, we present the experimental results, with additional results provided in Appendix~\ref{app:additional}. 

\textbf{KSE (E1).} This case is particularly challenging in the few-trajectory, coarse-resolution regime because of a stiff high-order operator. Physics-aware baselines struggle with these difficulties because of FD stencils, whose truncation errors become pronounced on coarse grids when approximating fourth-order derivatives. Data-driven FNO, while benefiting from spectral representations, learns a black-box mixing of Fourier coefficients and thus requires substantially more data to identify the correct operator behavior; under only a few trajectories, it tends to overfit and drift over time. In contrast, SINO explicitly learns stiff operators via Freq2Vec and models the nonlinearity through the $\Pi$-block. As shown in Table~\ref{tab:model_performance}, these structured and stability-aware design choices allow SINO to achieve the lowest error (\textbf{0.0040}) on KSE, outperforming the best data-driven baseline FNO ($0.0780$) and the best physics-aware baseline PeSANet ($0.2028$).

\begin{figure}
  \vspace{-1pt}
  \centering
  \includegraphics[width=0.8\linewidth]{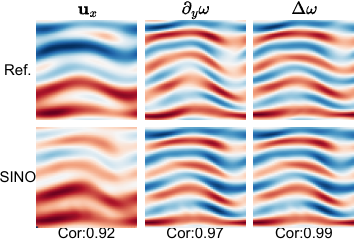} 
  \vspace{-2pt}
  \caption{\textbf{Feature maps learned by SINO.} Feature maps from $\Pi$-block and linear block align with ground-truth operators (e.g., $\partial_y\omega$, $\Delta\omega$) on E2.}
  \vspace{-6pt}
  \label{fig:operator_features}
\end{figure}

\textbf{NSE (E2--E5).} For the NSE cases, the challenge lies not only in capturing the nonlinear coupling between velocity and vorticity without explicit priors, but also in extrapolating beyond the training horizon. SINO maintains high correlation with the ground truth over long rollouts (Fig.~\ref{fig:corr}) and preserves coherent vortex structures even at the final timestep (Fig.~\ref{fig:mainresult}). By contrast, PeRCNN, P$^2$C$^2$Net, and PeSANet struggle under the coarse observation timestep (i.e., the subsampled effective $\Delta t$). Their FD-stencil-based operator representations are more sensitive to large effective $\Delta t$, leading to faster error accumulation and degraded rollouts. In comparison, SINO, inspired by spectral discretizations, provides a more faithful coarse-step evolution. Among purely data-driven baselines, FNO performs best, yet it still fails to resolve multiscale vortices. Quantitatively, SINO achieves $0.0049$--$0.0199$ relative error across E2--E5 and improves over the strongest baseline (FNO) by about $3.3\times$--$23.6\times$ depending on the setting (Table~\ref{tab:model_performance}). Furthermore, feature maps from SINO’s $\Pi$/linear blocks correlate with ground-truth NSE operators (e.g., $\partial_x\omega$, $\Delta\omega$), suggesting that SINO captures physically meaningful differential structure (Fig.~\ref{fig:operator_features}).

\textbf{Burgers (E6--E8).} 
For 2D Burgers (E6), physics-aware models with hard-coded local operator structure (e.g., PeRCNN/PeSANet) benefit from the equation’s local advection-diffusion form and thus remain competitive among baselines. However, SINO still achieves substantially higher accuracy, which reaches {0.0110} relative error, improving the best baseline P$^2$C$^2$Net ({0.0859}) by {7.8$\times$} (Table~\ref{tab:model_performance}). In the 3D case (E7), preserving consistent coupled structures across all three velocity components is markedly more challenging. Fig.~\ref{fig:3dmainresult} shows that SINO better retains small-scale shock and dissipation structures across $(u,v,w)$. Finally, in the mixed boundary-condition setting (E8), SINO remains robust with an error of {0.0376}, improving over the best data-driven baseline CNext ({0.1803}) by {4.8$\times$} and over the best physics-aware baseline P$^2$C$^2$Net ({0.1955}) by {5.2$\times$}. This indicates that SINO’s spectral-inspired representation is not restricted to fully periodic domains and can remain effective under non-periodic boundary constraints.

\begin{table}[t]
\centering
\caption{\textbf{Annular heat-conduction benchmark.} Relative $\ell_2$ error.}
\vspace{-2pt}
\setlength{\tabcolsep}{5pt}
\renewcommand{\arraystretch}{1.12}
\begin{tabular}{@{}lcccc@{}}
\toprule
\textbf{Metric}
& \textbf{Geo-FNO}~\citep{li2023Gfourier}
& \textbf{FFNO}~\citep{tran2023factorized}
& \textbf{LSM}~\citep{wu2023LSM}
& \textbf{SINO} \\
\midrule
Error
& 0.2804
& 0.2189
& \underline{0.1797}
& \textbf{0.0209} \\
\bottomrule
\end{tabular}
\vspace{-4pt}
\label{tab:annular_heat}
\end{table}

We further evaluate SINO on an annular heat-conduction benchmark to examine its performance under more general geometries and boundary conditions. The problem is defined on a non-Cartesian polar grid $(r,\theta)$ with $r\in[0.5,1.0]$, where Dirichlet boundary conditions are imposed along the radial direction. As shown in Table~\ref{tab:annular_heat}, SINO achieves a relative $\ell_2$ error of $0.0209$, substantially outperforming Geo-FNO ($0.2804$), FFNO ($0.2189$), and LSM ($0.1797$). This result indicates that the proposed spectral-inspired design remains effective beyond standard rectangular Cartesian domains, and can accommodate more complex geometries through boundary-aware spectral transforms.

\subsection{OOD Generalization to Different ICs}

\begin{figure*}[!t]
    \centering
    \includegraphics[width=1\linewidth]{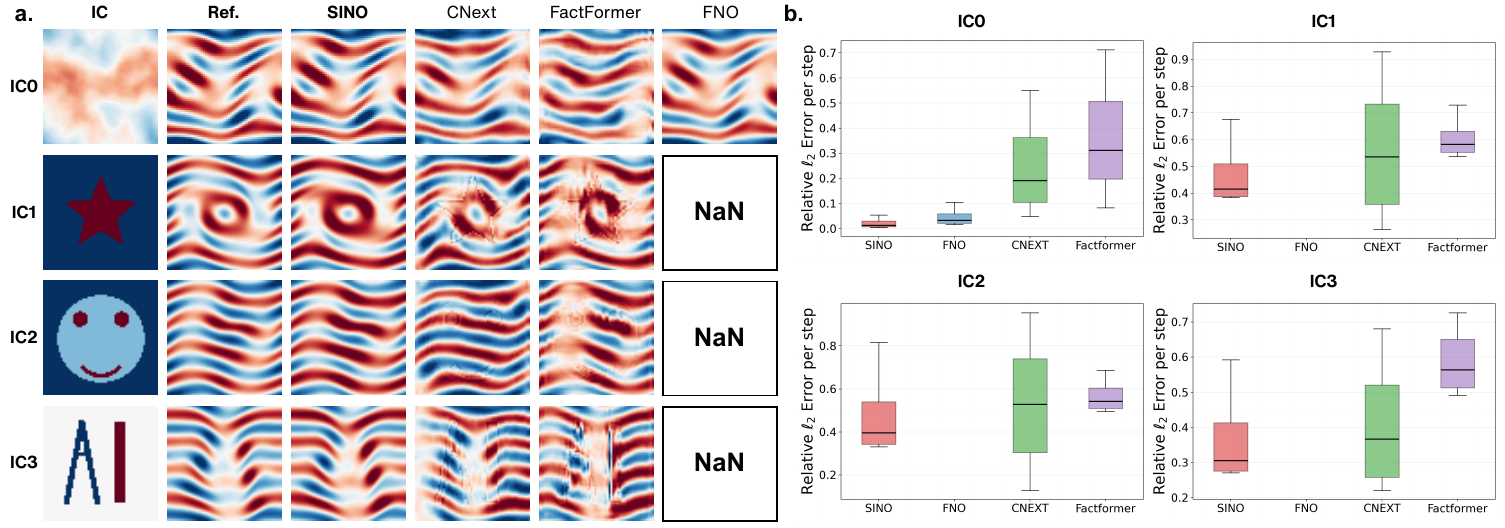}
    \vspace{-2pt}
\caption{\textbf{In-distribution vs.\ OOD generalization on NSE (E2).} (a) Vorticity-field visualizations and (b) boxplots comparing SINO trained with only \textbf{5} trajectories against data-driven baselines trained with \textbf{200} trajectories. IC0 is drawn from the training distribution, whereas IC1--IC3 are sampled from OOD initial-condition distributions. Results are shown at $t=15$\,s.}
    \label{fig:ood}
\end{figure*}

Generalizing to out-of-distribution (OOD) initial conditions (ICs) remains challenging for operator learning, especially when training data are limited. We evaluate this on the NSE task (E2) in a data-scarce setting: \textbf{SINO is trained with only 5 trajectories}, whereas data-driven baselines (CNext, FactFormer, FNO) are trained with \textbf{200} trajectories ($40\times$ more data). We test on the in-distribution case (IC0) and three OOD patterns: IC1 (Star), IC2 (Smiley face), and IC3 (``AI'' text). As shown in Fig.~\ref{fig:ood}a, all methods perform comparably on IC0, but their behavior diverges under OOD ICs. The baselines often produce severe artifacts or become numerically unstable (\texttt{NaN}) during long rollouts, failing to maintain coherent structures. In contrast, SINO remains stable and tracks the reference dynamics across all OOD patterns, preserving the advected shapes at $t=15$\,s. Fig.~\ref{fig:ood}b quantifies this gap via the per-step relative $\ell_2$ error over the rollout, where SINO consistently achieves the lowest median error on OOD cases. Overall, these results demonstrate that SINO does not rely on data memorization but learns the underlying operator mapping functions to functions. Additional rollout trajectories are provided in Appendix~\ref{app:rollout}.

\subsection{Ablation Studies}

\begin{table}
\centering
\caption{\textbf{Ablation results.} The relative $\ell_2$ error (\textbf{Error}) and high-correlation time (\textbf{HCT}, correlation $>0.8$; in seconds).}
\vspace{-2pt}
\label{tab:ablation}
\setlength{\tabcolsep}{3pt}
\resizebox{0.98\linewidth}{!}{
\begin{tabular}{l|cc|cc|cc}
\toprule
\multirow{2}{*}{\textbf{Model}} & \multicolumn{2}{c|}{\textbf{KSE}} & \multicolumn{2}{c|}{\textbf{NSE}} & \multicolumn{2}{c}{\textbf{Burgers}} \\
\cmidrule(lr){2-3} \cmidrule(lr){4-5} \cmidrule(lr){6-7}
& \textbf{Error} & \textbf{HCT} & \textbf{Error} & \textbf{HCT} & \textbf{Error} & \textbf{HCT} \\
\midrule
\textbf{SINO \textbackslash~$\Pi$-block} & 0.4730 & 2.3  & 0.8646 & 1.85 & 0.2290 & 2 \\
\textbf{SINO \textbackslash~Filter}      & 0.0131 & 5    & 0.1675 & 15   & 0.0124 & 2 \\
\textbf{SINO \textbackslash~Freq2Vec}    & 0.0057 & 5    & 0.0753 & 15   & 0.0634 & 2 \\
\textbf{SINO \textbackslash~Linear}      & 0.2882 & 5    & 0.0795 & 15   & 0.0915 & 2 \\
\textbf{SINO \textbackslash~RK4}         & 0.0623 & 5    & NaN    & 8.05 & 0.0144 & 2 \\
\midrule
\textbf{SINO}                             & \textbf{0.0040} & \textbf{5}  & \textbf{0.0199} & \textbf{15} & \textbf{0.0110} & \textbf{2} \\
\bottomrule
\end{tabular}}
\vspace{-2pt}
\end{table}

We conduct ablation experiments on three systems to evaluate the contributions of individual modules in Table~\ref{tab:ablation}. The experiments are conducted on three tasks: KSE (E1), NSE (E3), and Burgers (E6). Replacing the $\Pi$-block with a linear combination (\textbf{SINO \textbackslash~$\Pi$-block}) weakens overall performance, particularly for the NSE, which relies on crucial nonlinear terms: the error increases from $0.0199$ to $0.8646$, and the high-correlation time (HCT) drops sharply from $15$\,s to $1.85$\,s, indicating poor long-horizon stability. Removing the low-pass filter (\textbf{SINO \textbackslash~Filter}) also degrades rollout quality, with aliasing effects becoming most pronounced in the NSE system, where the error rises to $0.1675$ (vs.\ $0.0199$ for SINO). Here, \textbf{SINO \textbackslash~Freq2Vec} replaces Freq2Vec with an FNO-style learnable spectral weight matrix (i.e., per-mode mixing in the frequency domain). The clear drop verifies that Freq2Vec's shared mapping over frequency indices provides a non-trivial inductive bias, rather than a mere reparameterization of FNO kernels. Additionally, substituting the Freq2Vec module with a learnable vector (\textbf{SINO \textbackslash~Freq2Vec}) or removing the linear block (\textbf{SINO \textbackslash~Linear}) significantly degrades performance; for example, on NSE the error increases to $0.0753$ and $0.0795$, respectively, and on Burgers the error increases from $0.0110$ to $0.0634$ and $0.0915$. Replacing the RK4 scheme with an Euler integrator (\textbf{SINO \textbackslash~RK4}) reduces accuracy and long-term stability: the NSE rollout diverges (NaN), and HCT decreases to $8.05$\,s. These experiments demonstrate that every component of the SINO design is indispensable.

\begin{figure}
    \centering
    \includegraphics[width=1\linewidth]{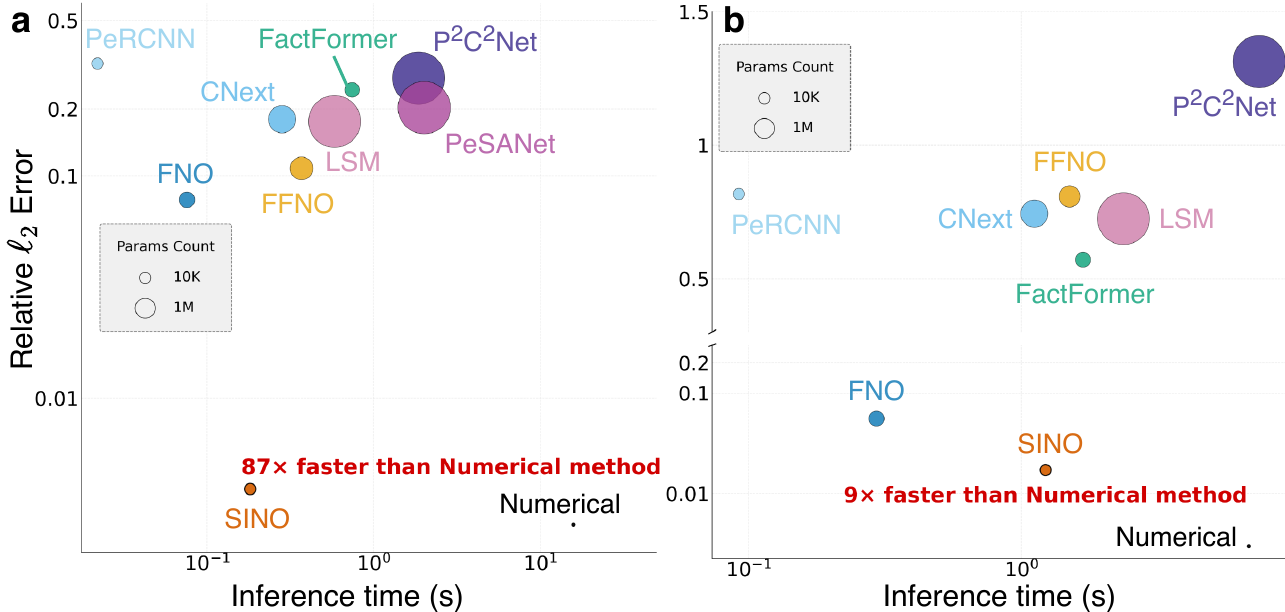}
    \caption{\textbf{Efficiency trade-off on (a) KSE and (b) NSE.} Bubble area scales with parameter count.}
    \label{fig:efficiency}
    \vspace{-4pt}
\end{figure}

\subsection{Inference Efficiency Analysis}
Fig.~\ref{fig:efficiency} compares inference time and accuracy on (a) KSE (E1) and (b) NSE (E2), with bubble size indicating parameter count. The numerical solver uses $\Delta t$ as the largest time step that maintains high accuracy and numerical stability at the given resolution.
SINO achieves the best accuracy while remaining fast at inference: it is \textbf{87$\times$} faster than the numerical solver on KSE and \textbf{9$\times$} faster on NSE (as annotated), yet retains low relative $\ell_2$ error. 
Several lightweight baselines (e.g., FNO) run slightly faster, but their errors are much higher (e.g., $0.0780$ vs.\ $0.0040$ on KSE). 
Conversely, some physics-aware models incur noticeably larger latency without closing the accuracy gap. 
Overall, SINO offers a strong accuracy--speed trade-off, providing substantial solver-level speedups without sacrificing predictive quality.

\subsection{Additional Numerical Results}

In Appendix~\ref{app:additional}, we provide further discussions and experiments beyond the main results. These include: (i) an analysis of how training data quantity affects SINO’s performance, (ii) parameter sensitivity studies that demonstrate SINO is robust to hyperparameter variations, (iii) evaluations of zero-shot discretization invariance, showing that SINO generalizes across unseen spatial resolutions without retraining, (iv) full-case snapshot visualizations and additional trajectory predictions of physical fields for more comprehensive comparison, (v) results of an RK4-trained FNO variant, showing that adopting RK4 alone does not yield comparable gains, and (vi) results of a small-capacity FNO, indicating that scaling FNO does not consistently close the performance gap to SINO.

\section{Conclusion}
\label{sec:conclusion}

We presented SINO, a spectral-inspired neural operator designed to model PDE-governed systems from limited data without requiring explicit knowledge of the governing equations. By leveraging frequency-domain embeddings through the proposed Freq2Vec module and incorporating multiplicative interactions via the $\Pi$-block with de-aliasing, SINO provides a compact yet expressive architecture that balances accuracy, generalization, and computational efficiency. Theoretically, we established its universal approximation ability for a broad class of PDE operators. Empirically, SINO consistently outperforms both data-driven and physics-encoded baselines, achieving substantial error reductions while maintaining robust OOD generalization. Moreover, SINO offers practical speed advantages over numerical solvers at inference, enabling fast rollout prediction for downstream use (e.g., repeated simulation and control), with speedups of up to $87\times$ in our benchmarks. Our current study mainly focuses on Cartesian grids, and a promising direction is to extend SINO beyond regular domains to more complex geometries via geometry-adapted basis transformation or coordinate transformations.

\section*{Limitations and Ethical Considerations}
SINO is currently limited to rectangular domains and does not support irregular geometries or unstructured meshes; extending it via coordinate mappings or geometry-adapted spectral bases is left for future work. This work focuses on developing a general neural operator framework (SINO) for learning PDE-governed dynamics from scarce data. Our study does not involve human subjects, sensitive personal data, or applications with immediate societal harm. 

\begin{acks}
The work is supported by the Beijing Natural Science Foundation (No. F261002) and the National Natural Science Foundation of China (No. 62276269 and No. 62506367). R.Z. acknowledges support from the China Postdoctoral Science Foundation under Grant Number 2025M771582 and the Postdoctoral Fellowship Program of CPSF under Grant Number GZB20250408. The full version is available at \url{http://arxiv.org/abs/2505.21573}.
\end{acks}

\bibliographystyle{ACM-Reference-Format}
\bibliography{sample-base}

\newpage
\appendix

\section*{Appendix}
\startcontents[appendix]

\section{Proof of Theorem~\ref{thm}}\label{app:proof}
\paragraph{Theorem (Approximation ability of SINO)} Let $\Omega=[0,1)^d$ be the $d$-torus and define
\[
\mathcal U =\{\mathbf{u}:\Omega\to\mathbb R^C  | \mathbf{u}\ \text{is periodic, bandlimited to }\|\mathbf{k}\|_\infty\le k_{\max},\ 
\|\mathbf{u}\|_{L^2}\le B \}.
\]
For any bounded Fourier multipliers $\psi_{j,i}:\mathcal K\to\mathbb C$ on
$\mathcal K:=\{\mathbf{k}\in\mathbb Z^d:\|\mathbf{k}\|_\infty\le k_{\max}\}$,
define the Fourier-multiplier operators
\[
\widehat{T_{\psi_{j,i}}\mathbf{u}}(\mathbf{k}) = \psi_{j,i}(\mathbf{k})\,\widehat{\mathbf{u}}(\mathbf{k}),\qquad \mathbf{k}\in\mathcal K.
\]
Consider the RHS operator
\[
\mathcal R(\mathbf{u}) \ :=\ \sum_{j=1}^J \ \prod_{i=1}^{n_j} T_{\psi_{j,i}}\mathbf{u},
\]
where the products are taken pointwise in $\mathbf{x}$. Then for any $\epsilon>0$, there exists a SINO model $\mathcal S_\theta$ such that
\[
\sup_{\mathbf{u}\in\mathcal U}\ \big\|\mathcal S_\theta(\mathbf{u})-\mathcal R(\mathbf{u})\big\|_{L^2} \ <\ \epsilon.
\]

\begin{proof}
Fix $\epsilon>0$. Since $\mathcal K=\{\mathbf{k}\in\mathbb Z^d:\|\mathbf{k}\|_\infty\le k_{\max}\}$ is finite,
for each bounded multiplier $\psi_{j,i}:\mathcal K\to\mathbb C$ and any $\delta>0$, the shared MLP in Freq2Vec
can produce $\tilde\psi_{j,i}$ such that
\begin{equation}\label{eq:psi_delta}
\max_{\mathbf{k}\in\mathcal K}\,|\tilde\psi_{j,i}(\mathbf{k})-\psi_{j,i}(\mathbf{k})|\le \delta .
\end{equation}
Let $T_{\tilde\psi_{j,i}}$ be the corresponding approximate multiplier operators, and define
\[
\tilde{\mathcal R}(\mathbf{u}) := \sum_{j=1}^J \prod_{i=1}^{n_j} T_{\tilde\psi_{j,i}}\mathbf{u}.
\]

\paragraph{Step 1: Error of one multiplier operator.}
For any $\mathbf{u}\in\mathcal U$, by Parseval's identity and \eqref{eq:psi_delta},
\[
\|T_{\tilde\psi}\mathbf{u}-T_{\psi}\mathbf{u}\|_{L^2}^2
=\sum_{\mathbf{k}\in\mathcal K}|\tilde\psi(\mathbf{k})-\psi(\mathbf{k})|^2\,\|\hat{\mathbf{u}}(\mathbf{k})\|_2^2
\le \delta^2 \|\mathbf{u}\|_{L^2}^2
\le \delta^2 B^2,
\]
hence
\begin{equation}\label{eq:single_err}
\|T_{\tilde\psi}\mathbf{u}-T_{\psi}\mathbf{u}\|_{L^2}\le \delta B.
\end{equation}

\paragraph{Step 2: $L^\infty$ control for bandlimited factors.}
For any bandlimited field $\mathbf{v}(\mathbf{x})=\sum_{\mathbf{k}\in\mathcal K}\hat{\mathbf{v}}(\mathbf{k})e^{2\pi i \mathbf{k}\cdot\mathbf{x}}$,
we have, for all $\mathbf{x}$,
\[
\|\mathbf{v}(\mathbf{x})\|_2\le \sum_{\mathbf{k}\in\mathcal K}\|\hat{\mathbf{v}}(\mathbf{k})\|_2
\le \sqrt{|\mathcal K|}\Big(\sum_{\mathbf{k}\in\mathcal K}\|\hat{\mathbf{v}}(\mathbf{k})\|_2^2\Big)^{1/2}
= \sqrt{|\mathcal K|}\,\|\mathbf{v}\|_{L^2},
\]
so
\begin{equation}\label{eq:linfty_bd}
\|\mathbf{v}\|_{L^\infty}\le \sqrt{|\mathcal K|}\,\|\mathbf{v}\|_{L^2}.
\end{equation}
Moreover, for any multiplier $\psi$,
\[
\|T_\psi \mathbf{u}\|_{L^2}^2
=\sum_{\mathbf{k}\in\mathcal K}|\psi(\mathbf{k})|^2\|\hat{\mathbf{u}}(\mathbf{k})\|_2^2
\le \|\psi\|_\infty^2\|\mathbf{u}\|_{L^2}^2
\le \|\psi\|_\infty^2 B^2,
\]
hence $\|T_\psi\mathbf{u}\|_{L^\infty}\le \sqrt{|\mathcal K|}\,\|\psi\|_\infty B$ by \eqref{eq:linfty_bd}.
Let
\[
M_\delta:=\sqrt{|\mathcal K|}\,B\cdot \max_{j,i}(\|\psi_{j,i}\|_\infty+\delta),
\]
then for all $j,i$ and $\mathbf{u}\in\mathcal U$,
\begin{equation}\label{eq:factor_bd}
\|T_{\psi_{j,i}}\mathbf{u}\|_{L^\infty}\le M_\delta,\qquad
\|T_{\tilde\psi_{j,i}}\mathbf{u}\|_{L^\infty}\le M_\delta.
\end{equation}

\paragraph{Step 3: Error propagation through products.}
Fix $j$ and define $a_i:=T_{\tilde\psi_{j,i}}\mathbf{u}$, $b_i:=T_{\psi_{j,i}}\mathbf{u}$.
Using the telescoping identity,
\[
\prod_{i=1}^{n_j}a_i-\prod_{i=1}^{n_j}b_i
=\sum_{m=1}^{n_j}\Big(\prod_{i<m}a_i\Big)(a_m-b_m)\Big(\prod_{i>m}b_i\Big),
\]
and $\|fg\|_{L^2}\le \|f\|_{L^\infty}\|g\|_{L^2}$, together with \eqref{eq:single_err} and \eqref{eq:factor_bd},
we get
\[
\Big\|\prod_{i=1}^{n_j}a_i-\prod_{i=1}^{n_j}b_i\Big\|_{L^2}
\le \sum_{m=1}^{n_j} (M_\delta)^{n_j-1}\,\|a_m-b_m\|_{L^2}
\le n_j (M_\delta)^{n_j-1}\,\delta B.
\]

\paragraph{Step 4: Sum over $j$ and choose $\delta$.}
Summing over $j=1,\dots,J$ yields, uniformly for all $\mathbf{u}\in\mathcal U$,
\[
\|\tilde{\mathcal R}(\mathbf{u})-\mathcal R(\mathbf{u})\|_{L^2}
\le \delta B \sum_{j=1}^J n_j (M_\delta)^{n_j-1}
=: C_\delta\,\delta.
\]
Choose $\delta>0$ such that $C_\delta\,\delta\le \epsilon$ (e.g., take $\delta$ sufficiently small).
Then
\[
\sup_{\mathbf{u}\in\mathcal U}\ \|\tilde{\mathcal R}(\mathbf{u})-\mathcal R(\mathbf{u})\|_{L^2} < \epsilon.
\]

\paragraph{Step 5: Realizability by SINO.}
SINO's SLB implements each $T_{\tilde\psi_{j,i}}$ by FFT $\rightarrow$ element-wise multiplication with $\tilde\psi_{j,i}(\mathbf{k})$
generated by Freq2Vec $\rightarrow$ inverse FFT. The $\Pi$-block computes pointwise products of selected channels,
and a $1\times 1$ convolution linearly recombines the resulting product channels.
Hence there exists $\theta$ such that $\mathcal S_\theta(\mathbf{u})\equiv \tilde{\mathcal R}(\mathbf{u})$,
which completes the proof.
\end{proof}

\paragraph{On unbounded Fourier bandwidth.} The theorem assumes that the input function $\mathbf{u}$ is bandlimited with $\|\mathbf{k}\|_\infty \leq k_{\max}$. In practice, many physical fields are not exactly bandlimited. However, if $\mathbf{u} \in C^m(\Omega)$, then its Fourier coefficients decay at the rate
\[
|\hat{\mathbf{u}}(\mathbf{k})| \lesssim \frac{1}{\|\mathbf{k}\|^m},
\]
which implies that the tail $\|\mathbf{u} - \mathbf{u}_{k_{\max}}\|_{L^2}$ decays as $O(k_{\max}^{-(m - d/2)})$. Thus, the total approximation error of SINO in this case can be bounded by
\[
\| \mathcal{S}_\theta(\mathbf{u}) - \mathcal{R}(\mathbf{u}) \|_{L^2} \leq \epsilon + C_2 k_{\max}^{-(m - d/2)}.
\]

By choosing a sufficiently large $k_{\max}$ and network capacity, this bound can still be made arbitrarily small. 

\paragraph{Supported PDE types.} The class of PDEs covered by this theorem includes a broad family of physical systems, as long as the RHS can be expressed as a sum of products of linear differential operators. This encompasses reaction-diffusion equations, convection-diffusion equations, the incompressible Navier-Stokes equations, and other general polynomial-type PDEs.

\section{Experimental Details}

\subsection{Data Generation}\label{app:data}

Initial conditions for all PDE systems were sampled from a Gaussian Random Field (GRF), following the standard practice introduced in the classical FNO work~\citep{li2020fourier}. We target the data-scarce scenario: each experiment includes only 5 trajectories in the training set, as summarized in Table~\ref{tab:dataset_summary}. The table specifies the numerical scheme used for data generation, the spatial and temporal domains, and the number of training and testing trajectories for each PDE case. For the NSE case, training trajectories span 10\,s, while evaluation involves rollout prediction over 15\,s, as indicated in the temporal domain row of Table~\ref{tab:dataset_summary}. Numerical solutions were obtained using a $2/3$ dealiased pseudo-spectral method with periodic boundary conditions, implemented in Python with PyTorch, consistent with prior classical papers~\citep{li2020fourier,tran2023factorized}. In particular, the NSE dataset was generated using an open-source implementation from~\citep{li2020fourier}. To ensure reproducibility, we fixed random seeds for dataset splitting: seed 0 for training, seed 1 for validation, and seed 2 for testing.

\subsection{Baseline Models}\label{Baseline Models}

We provide detailed introductions to the baseline models used for comparison with SINO. These baselines include data-driven approaches such as FNO~\citep{li2020fourier}, FFNO~\citep{tran2023factorized}, FactFormer~\citep{li2024scalable}, CNext~\citep{liu2022convnet,ohana2024well} and LSM~\citep{wu2023LSM}, as well as physics-encoded models including PeRCNN~\citep{Rao_2023}, P$^2$C$^2$Net~\citep{WangEtAl2024NeurIPS} and PeSANet~\citep{wan2025pesanet}. Details of baseline models are provided as follows:

\textbf{Fourier Neural Operator (FNO)}~\citep{li2020fourier}. FNO is one of the most classical data-driven neural operator model that captures features in the frequency domain. Its ability to capture information in the frequency domain allows it to effectively utilize global information, and it also possesses a degree of resolution invariance. 

For hyperparameter tuning, we conducted a grid search over the Fourier modes ({8, 12, 16}), network width ({32, 64}), number of layers ({1, 4}), and learning rate ({0.01, 0.005, 0.001}), and selected the best configuration according to validation-set performance.

\textbf{Factorized Fourier Neural Operator (FFNO)}~\citep{tran2023factorized}. FFNO is an improved neural operator model based on FNO, and its core involves the introduction of factorized Fourier representation. This factorization method and improved network structure allow FFNO to employ deeper network architectures and demonstrate performance superior to standard FNO in simulating various partial differential equations.

For FFNO, we performed a grid search over the network width ({32, 64}), Fourier modes ({8, 12, 16}), and learning rate ({0.01, 0.005, 0.001}), and selected the best configuration based on validation-set performance.

\textbf{FactFormer}~\citep{li2024scalable}. FactFormer is a transformer-based model designed for multi-dimensional settings that leverages an axial factorized kernel integral, implemented via a learnable projection operator that decomposes the input function into one-dimensional sub-functions. 

For hyperparameter tuning, we searched over the hidden dimension ({64, 128, 256}), attention head dimension ({8, 16}), network depth ({2, 4}), and learning rate ({0.01, 0.005, 0.001}), and selected the best configuration according to validation-set performance.

\textbf{CNext}~\citep{liu2022convnet,ohana2024well}. CNext is a family of modernized ConvNet models, inspired by the design principles of Vision Transformers (ViTs) while retaining the efficiency and simplicity of convolutional architectures. In the latest benchmark study~\citep{ohana2024well}, CNext achieved SOTA results across a wide range of PDE learning tasks.

For hyperparameter tuning, we conducted a grid search over the network width ({16, 32, 64}) and number of layers ({2, 3, 4}), and learning rate ({0.01, 0.005, 0.001}), and selected the best configuration according to validation-set performance.

\textbf{LSM (Latent Spectral Models).}~\citep{wu2023LSM}.
LSM is a neural-operator framework designed for high-dimensional PDEs by moving the learning and solving process from the coordinate space to a compact latent space. 
It first uses an attention-based hierarchical projection network to compress high-dimensional fields into latent representations in linear time, and then applies a neural spectral block inspired by classical spectral methods to model PDE dynamics via learning multiple basis operators. 

For LSM, we conducted a grid search over the model width ({16, 32, 64}), patch size ({3, 5}), and learning rate ({0.01, 0.005, 0.001}), and selected the best hyperparameter configuration based on validation-set performance.

\textbf{Physics-embedded Recurrent-Convolutional Neural Network (PeRCNN)}~\citep{Rao_2023}. PeRCNN is a physics-encoded learning methodology that directly embeds physical laws into the neural network architecture. It employs multiple parallel convolutional neural network and leverages feature map multiplication to simulate polynomial equations, thereby enhancing the model's extrapolation and generalization capabilities. 

For hyperparameter tuning, we performed a grid search over the network width ({32, 64, 128}) and input kernel size ({3, 5, 7}),  and learning rate ({0.01, 0.005, 0.001}), and selected the best hyperparameter configuration based on validation-set performance.

\textbf{Physics-encoded Spectral Attention Network (PeSANet)}~\citep{wan2025pesanet}. PeSANet includes a physics-encoded block for approximating local differential operators and a spectral-enhanced block which, combined with spectral attention, captures global features in the frequency domain, allowing it to perform excellently, especially in long-term forecasting accuracy, under scarce data and incomplete physical priors. 

For hyperparameter tuning, we searched over the network width ({32, 64, 128}) and input kernel size ({5, 7}), and learning rate ({0.01, 0.005, 0.001}), and selected the best hyperparameter configuration based on validation-set performance.

\textbf{PDE-Preserved Coarse Correction Network (P$^2$C$^2$Net)}~\citep{WangEtAl2024NeurIPS}. 
P$^2$C$^2$Net is a physics-encoded model designed to solve spatiotemporal PDEs efficiently on coarse grids under small-data regimes. 
It consists of two complementary modules: (i) a trainable PDE block that updates the coarse system state based on a high-order numerical scheme with boundary-condition encoding, and (ii) a neural correction block that refines the solution online during rollout. 

For P2C2Net, we conducted a grid search over hidden channels ({8, 16, 32}), input kernel size ({3, 5}), and learning rate ({0.01, 0.005, 0.001}), and selected the best configuration according to validation-set performance.

For the proposed \textbf{SINO}, we conducted a grid search over the number of channels ({32, 64}) and the dimension of $\psi(\mathbf{k})$ in the Freq2Vec ({6, 8, 12}), and learning rate ({0.01, 0.005, 0.001}), and selected the best hyperparameter configuration based on validation-set performance.

Our model follows standard machine learning practices, where all hyperparameters are tuned based on validation performance with fixed seed, ensuring fair and consistent model selection. The parameter sizes of all baseline models and SINO are summarized in Table~\ref{tab:model_parameternum}. Table~\ref{tab:model_parameternum} highlights a striking gap in parameter budgets: SINO operates in the $10^3$-parameter range, whereas many neural spectral baselines are typically $10^5$--$10^8$ on the same tasks. This compactness is not obtained by arbitrarily shrinking the network, but comes from how SINO represents frequency-domain operators. In particular, Freq2Vec parameterizes the spectral multiplier as a shared function of the frequency index,
\[
m(\mathbf{k})=\psi_{\theta}(\mathbf{k}),
\]
so the same small set of parameters is reused across all modes. This matches a basic property of PDE operators, where derivative multipliers are structured functions of $\mathbf{k}$ (e.g., polynomials in $|\mathbf{k}|$), and it serves as an effective form of capacity control in the few-shot regime.

By contrast, many Fourier-based layers assign independent learnable weights to retained modes (for example, per-mode mixing matrices in FNO), so the number of free parameters grows quickly with the number of modes and the dimensionality. In low-data settings, simply reducing a baseline's parameters by cutting channels or truncating modes often hurts for a concrete reason: fewer channels weaken feature interactions, and fewer modes remove high-frequency content that is essential for sharp structures and stable rollouts. SINO sidesteps this trade-off by keeping the mode set intact while constraining the hypothesis class through the shared, continuous mapping over $\mathbf{k}$, which preserves frequency coverage without paying a large parameter cost.

\begin{table*}[!t]
    \centering
    \caption{\textbf{Parameter counts of different models.} NA denotes `Not Applicable'. $k$ and $M$ denote $10^3$ and $10^6$ parameters, respectively.}
    \vspace{3pt}
    \resizebox{15.0cm}{!}{
    \begin{tabular}{l|c|cccc|ccc}
        \toprule
        \multirow{2}{*}{\makecell{\ \\ \textbf{Model}\\ \ }}
        & \textbf{KSE} 
        & \multicolumn{4}{c|}{\textbf{NSE}} 
        & \multicolumn{3}{c}{\textbf{Burgers}} \\
        \cmidrule(lr){2-2} \cmidrule(lr){3-6} \cmidrule(lr){7-9}
        & \makecell{E1\\ \ } 
        & \makecell{E2\\$10^{-4}, f_1$} 
        & \makecell{E3\\$10^{-5}, f_1$} 
        & \makecell{E4\\$10^{-4}, f_2$} 
        & \makecell{E5\\$10^{-5}, f_2$} 
        & \makecell{E6\\2D}  
        & \makecell{E7\\3D}  
        & \makecell{E8\\2D-MixedBC} \\
        \midrule
        \textbf{FNO}
        & 0.30M & 0.30M & 0.30M & 1.19M & 4.74M & 1.19M & 127.41M & 1.19M \\

        \textbf{FFNO}
        & 1.68M & 1.29M & 0.32M & 0.89M & 0.89M & 1.68M & 62.4k & 1.68M \\

        \textbf{FactFormer}
        & 0.19M & 0.28M & 0.19M & 0.19M & 0.28M & 0.19M & 0.30M & 0.31M \\

        \textbf{CNext}
        & 3.15M & 3.15M & 3.15M & 3.15M & 3.15M & 3.15M & 4.16M & 51.35M \\

        \textbf{LSM}
        & 19.20M & 19.20M & 4.82M & 19.20M & 4.82M & 19.20M & NA & 4.82M \\

        \textbf{P$^2$C$^2$Net}
        & 19.47M & 19.47M & 19.47M & 19.47M & 19.47M & 19.47M & NA & 19.47M \\

        \textbf{PeSANet}
        & 19.54M & 19.54M & 19.54M & 19.54M & 19.54M & 19.54M & NA & 19.54M \\

        \textbf{PeRCNN}
        & 10.1k & 10.1k & 10.1k & 10.1k & 10.1k & 5.1k & 1.9k & 5.1k \\
        \midrule
        \textbf{SINO (ours)}
        & \textbf{8.4k} & \textbf{3.4k} & \textbf{2.7k} & \textbf{6.8k} & \textbf{6.8k} & \textbf{9.2k} & \textbf{3.0k} & \textbf{13.9k} \\
        \bottomrule
    \end{tabular}}
    \label{tab:model_parameternum}
\end{table*}

\subsection{Metrics}\label{Metrics}
To assess the performance of our proposed method, we utilize three evaluation metrics: relative $\ell_2$ error, correlation, and High-Correlation Time (HCT). The relative $\ell_2$ error measures the ratio of the $\ell_2$ norm of the error vector to that of the ground-truth vector, providing a dimensionless measure of the prediction error relative to the true scale. Correlation is quantified using the Pearson correlation coefficient (PCC), which measures the linear dependence between predicted and ground-truth solutions over time. High-Correlation Time (HCT) measures the duration (in seconds) during which the model maintains a high correlation with the ground truth along a rollout.

The definitions are as follows:
\begin{equation}
\label{metric}
\begin{aligned}
\text{Relative } \ell_2 \text{ Error:}& \quad 
\frac{\|\mathbf{y} - \mathbf{\tilde{y}}\|_2}{\|\mathbf{y}\|_2} 
= \sqrt{\frac{\sum_{i=1}^{n} (y_i - \tilde{y}_i)^2}{\sum_{i=1}^{n} y_i^2}}, \\
\text{Correlation (PCC):} & \quad 
\text{PCC}(\mathbf{y}, \mathbf{\tilde{y}}) 
= \frac{\text{Cov}(\mathbf{y}, \mathbf{\tilde{y}})}{\sigma_{\mathbf{y}} \sigma_{\mathbf{\tilde{y}}}}, \\
\text{HCT:} & \quad
\text{HCT} = \sum_{t=1}^{T} \Delta t \cdot \mathbb{I}\!\left(\text{PCC}(\mathbf{y}_t, \mathbf{\tilde{y}}_t) > 0.8\right).
\end{aligned}
\end{equation}
Here, $n$ denotes the number of evaluation points, $y_i$ and $\tilde{y}_i$ are the $i$-th components of the ground-truth vector $\mathbf{y}$ and the predicted vector $\mathbf{\tilde{y}}$, respectively, $\text{Cov}$ is the covariance, and $\sigma$ denotes the standard deviation. For HCT, $\mathbf{y}_t$ and $\mathbf{\tilde{y}}_t$ denote the ground-truth and predicted solutions at timestep $t$, $\Delta t$ is the rollout timestep, $T$ is the total number of rollout steps, and $\mathbb{I}(\cdot)$ is the indicator function.

\subsection{Training Details}\label{Training Details}

All experiments were conducted on a single NVIDIA A100 GPU (80GB) with an Intel(R) Xeon(R) Platinum 8380 CPU (2.30GHz, 64 cores). For 2D cases, models were trained for 20000 iterations, while for 3D cases, 5000 iterations were used. We adopt a OneCycle learning rate scheduler with a maximum learning rate selected from $\{0.01,0.001\}$, scheduled over the entire training horizon.

During training, we randomly sample a starting position along each trajectory. From this state, the model first evolves forward $n$ steps without gradients (warm-up), where $n$ is randomly sampled from $\{0,\ldots,n_1\}$ at each iteration, and then predicts the following $n_2$ steps with gradients, which are compared against the ground truth. We fix $n_1=4$ and $n_2=8$. This scheme provides stable backpropagation and encourages the models to generalize across different rollout horizons~\citep{brandstetter2022message}. 

\section{Additional Numerical Results}\label{app:additional}

\subsection{Data Efficiency}

To evaluate the data efficiency of SINO, we vary the number of training trajectories and report the results in Table~\ref{tab:data_efficiency}. Two observations stand out. First, SINO remains accurate in the low-data regime. With only \textbf{5} trajectories, it already achieves a relative $\ell_2$ error of \textbf{0.0040} on KSE and \textbf{0.01710} on NSE, indicating that SINO can learn meaningful operator structure from a small number of trajectories. 
Second, additional data provides diminishing but consistent gains. Increasing the training set from 5 to 15 trajectories improves NSE error from 0.01710 to \textbf{0.00939}, while KSE saturates early (0.0040 at 5 trajectories and 0.0039 at 10--15 trajectories). 
Notably, the NSE setting is more data-demanding: training with only \textbf{2} trajectories leads to a sharp degradation (0.77720), whereas KSE remains relatively stable (0.0065). Overall, these results show that SINO is data-efficient, while the degree of data sensitivity depends on the underlying dynamics.

\begin{table}[!t]
\centering
\caption{\textbf{Data efficiency.} Relative $\ell_2$ error vs. number of training trajectories.}
\vspace{-2pt}
\resizebox{0.5\linewidth}{!}{
\begin{tabular}{c|cc}
\toprule
\textbf{\#Traj.} & \textbf{NSE} & \textbf{KSE} \\
\midrule
15 & 0.00939 & 0.0039 \\
10 & 0.01026 & 0.0039 \\
5  & 0.01710 & 0.0040 \\
2  & 0.77720 & 0.0065 \\
\bottomrule
\end{tabular}}
\vspace{-2pt}
\label{tab:data_efficiency}
\end{table}

\subsection{Hyperparameter Sensitivity Analysis}
Table~\ref{tab:sino_ns_hparam_e4} reports the sensitivity of SINO to two key hyperparameters on NSE (E4): the channel size ($C \in \{64, 32\}$) and the embedding dimension $K$ of $\psi(\mathbf{k})$ in the Freq2Vec module ($K \in \{12, 8, 6\}$). Overall, SINO remains stable across a reasonable range of settings, with validation relative $\ell_2$ errors between 0.0049 and 0.0175. The best result is achieved at $(K=6, C=64)$ with an error of \textbf{0.0049}. This indicates that SINO is not overly sensitive to moderate hyperparameter changes.

\begin{table}[!ht]
\centering
\caption{\textbf{Hyperparameter sensitivity of SINO on NSE (E4).} Best validation relative $\ell_2$ error for each $(K,C)$ setting.}
\label{tab:sino_ns_hparam_e4}
\begin{tabular}{ccc}
\toprule
Model & $C=64$ & $C=32$ \\
\midrule
$K=12$ & 0.0148 & 0.0055 \\
$K=8$  & 0.0059 & 0.0155 \\
$K=6$  & \textbf{0.0049} & 0.0175 \\
\bottomrule
\end{tabular}
\end{table}

\subsection{Zero-shot Super-resolution}

We further evaluate the zero-shot resolution transfer of SINO. Specifically, the model is trained at a fixed resolution of $64\times64$ and then directly evaluated at other resolutions without retraining. As summarized in Table~\ref{tab:resolution_sensitivity}, SINO generalizes reasonably well across grid resolutions: when evaluated at the training resolution ($64\times64$), the relative $\ell_2$ error is $0.0103$, and it further decreases to $0.0086$ on a finer grid ($128\times128$). In contrast, the error increases on coarser grids due to reduced spatial fidelity, rising to $0.0363$ at $32\times32$ and $0.2935$ at $16\times16$. Overall, these results indicate that SINO can be applied in a zero-shot manner across resolutions, and it can benefit from higher-resolution inputs at inference time while degrading gracefully when the resolution is reduced.

\begin{table}[!t]
\centering
\caption{\textbf{Zero-shot resolution transfer.} Relative $\ell_2$ error under different spatial resolutions (trained on $64\times64$ and evaluated on other resolutions without retraining).}

\vspace{-2pt}
\begin{tabular}{c|c}
\toprule
\textbf{Resolution} & \textbf{Relative error (mean)} \\
\midrule
$128\times128$ & 0.0086 \\
$64\times64$   & 0.0103 \\
$32\times32$   & 0.0363 \\
$16\times16$   & 0.2935 \\
\bottomrule
\end{tabular}
\vspace{-2pt}
\label{tab:resolution_sensitivity}
\end{table}

\begin{table}[!ht]
\centering
\caption{Relative $\ell_2$ error of replacing one-step rollout with RK4 in FNO.}
\label{tab:fno_rk4}
\begin{tabular}{lccc}
\toprule
Case & FNO-RK4 & FNO & SINO \\
\midrule
E1 & 0.0961 & 0.0780 & 0.0040 \\
E2 & 0.0728 & 0.0558 & 0.0171 \\
E6 & 0.3248 & 0.3616 & 0.0110 \\
\bottomrule
\end{tabular}
\end{table}

\subsection{Comparison with FNO-RK4}
\label{subsec:fno_rk4}

To isolate the effect of the RK4 integrator, we construct an RK4 variant of FNO (denoted as \textbf{FNO-RK4}) and compare it with the standard one-step FNO rollout. \textbf{FNO-RK4} is trained to represent a continuous-time vector field: given a state $u_t$, the network predicts a time increment (equivalently, an approximation of $\Delta t\, f(u_t)$), and the trajectory is advanced using the classical fourth-order Runge--Kutta scheme. 
We tune \textbf{FNO-RK4} with a validation-based grid search using the same hyperparameter ranges and a comparable training budget as FNO (see Appendix~\ref{Baseline Models} for details). 
As shown in Table~\ref{tab:fno_rk4}, using RK4 in FNO yields performance comparable to (and sometimes worse than) the standard FNO, while still remaining far behind SINO. This indicates that the improvements of SINO cannot be attributed to the RK4 integrator alone, but rather stem from its spectral-structured operator parameterization and stability-aware design.

\begin{table}[t]
\centering
\caption{Performance of small-capacity FNO variants of E6.}
\label{tab:small_fno}
\begin{tabular}{rcc}
\toprule
Width $w$ & \# Parameters & Mean error \\
\midrule
2 & 1{,}810  & 0.7902 \\
4 & 5{,}546  & 0.3822 \\
8 & 19{,}954 & 0.4764 \\
\bottomrule
\end{tabular}
\end{table}

\subsection{The Results of Small FNO}
To examine whether a compact data-driven baseline can match SINO in the data-scarce setting, we train small FNO models by restricting the architecture to a single FNO layer and varying the channel width ($w\in\{2,4,8\}$).
As summarized in Table~\ref{tab:small_fno}, increasing the width from $2$ to $4$ reduces the error, but further increasing to $8$ does not yield additional gains in this regime.
These results suggest that simply scaling the capacity of a data-driven FNO baseline (within this small-parameter range) is insufficient to close the performance gap to SINO under our data-scarce coarse-grid setting.

\subsection{Additional Rollout Trajectories}
\label{app:rollout}

\begin{figure*}
    \centering
    \includegraphics[width=1\linewidth]{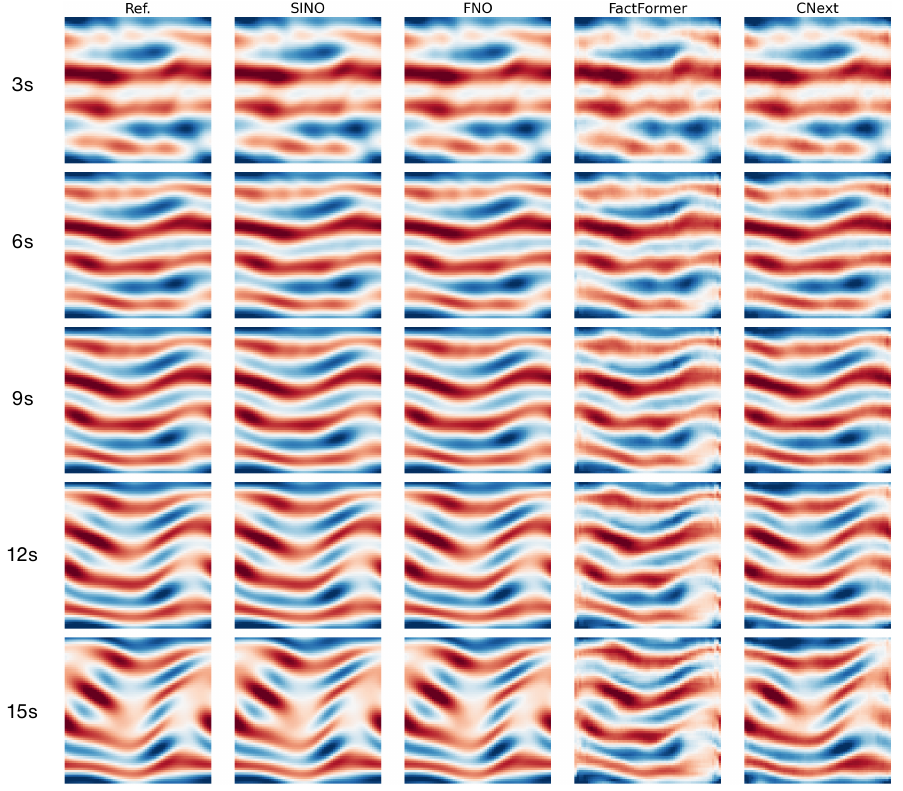}
    \caption{\textbf{In-distribution trajectories on IC0.} Comparison between SINO trained with only \textbf{5} trajectories and data-driven baselines trained with \textbf{200} trajectories. }
    \label{fig:IC0tra}
\end{figure*}

\begin{figure*}
    \centering
    \includegraphics[width=1\linewidth]{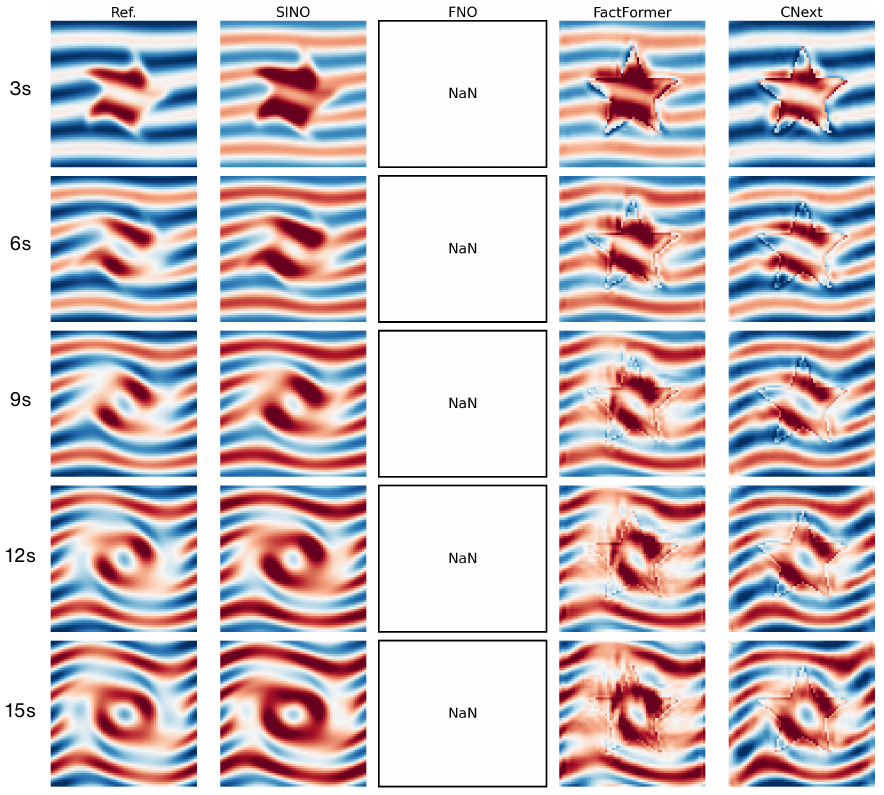}
    \caption{\textbf{OOD generalization trajectories on IC1 (star).} Comparison between SINO trained with only \textbf{5} trajectories and data-driven baselines trained with \textbf{200} trajectories. }
    \label{fig:IC1tra}
\end{figure*}

\begin{figure*}
    \centering
    \includegraphics[width=1\linewidth]{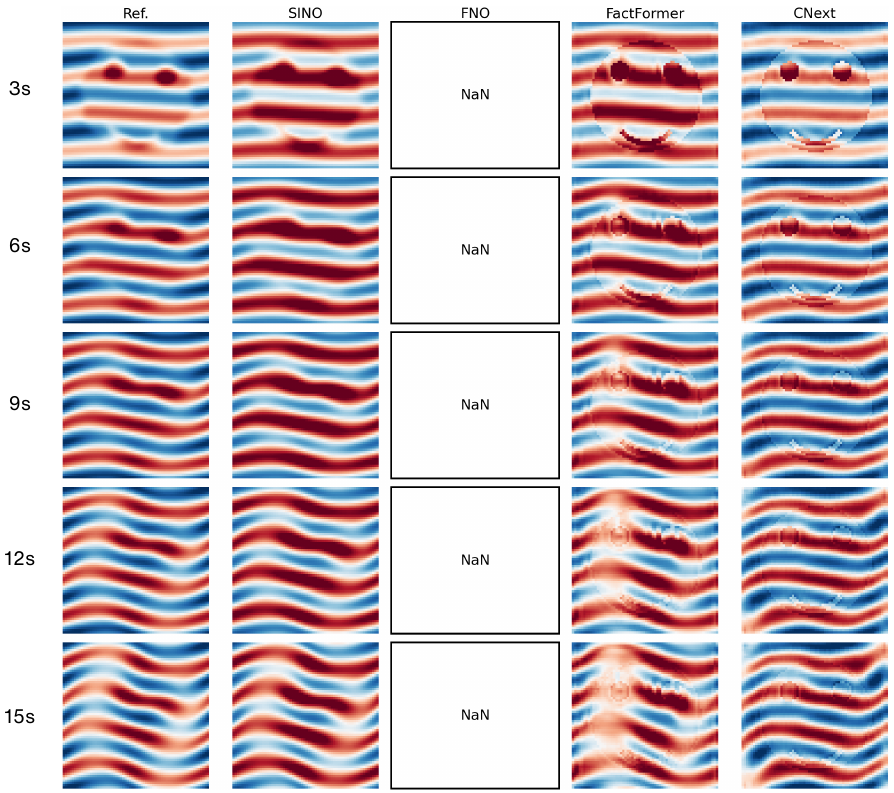}
    \caption{\textbf{OOD generalization trajectories on IC2 (smiley face).} Comparison between SINO trained with only \textbf{5} trajectories and data-driven baselines trained with \textbf{200} trajectories.}
    \label{fig:IC2tra}
\end{figure*}

\begin{figure*}
    \centering
    \includegraphics[width=1\linewidth]{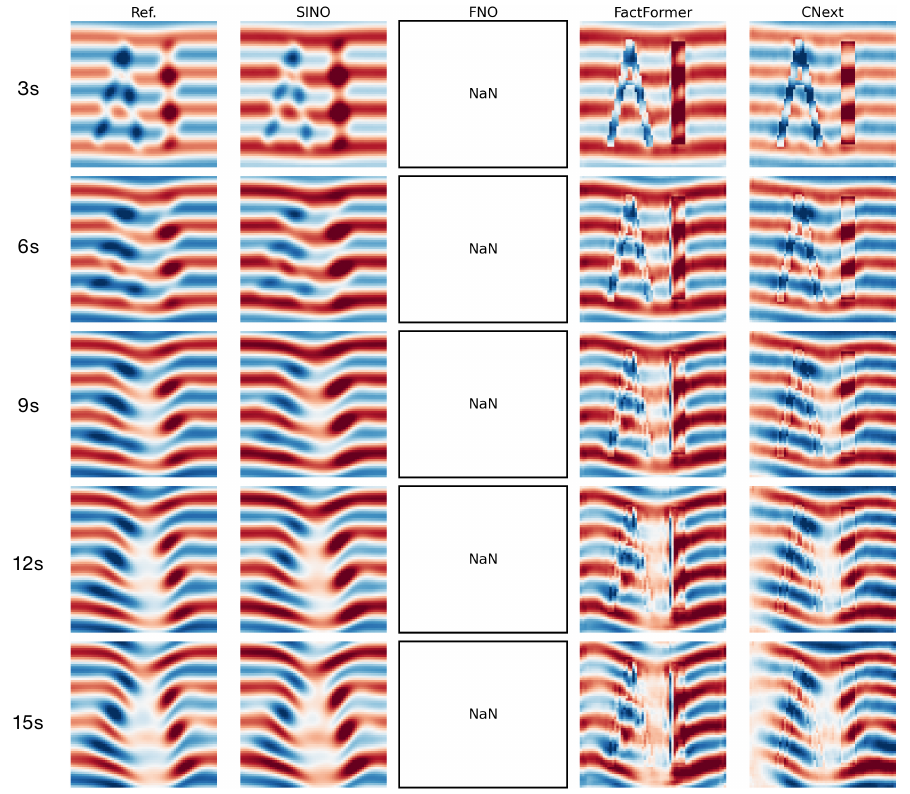}
    \caption{\textbf{OOD generalization trajectories on IC3 (the pattern ‘AI’).} Comparison between SINO trained with only \textbf{5} trajectories and data-driven baselines trained with \textbf{200} trajectories.}
    \label{fig:IC3tra}
\end{figure*}

\begin{figure*}
    \centering
    \includegraphics[width=0.8\linewidth]{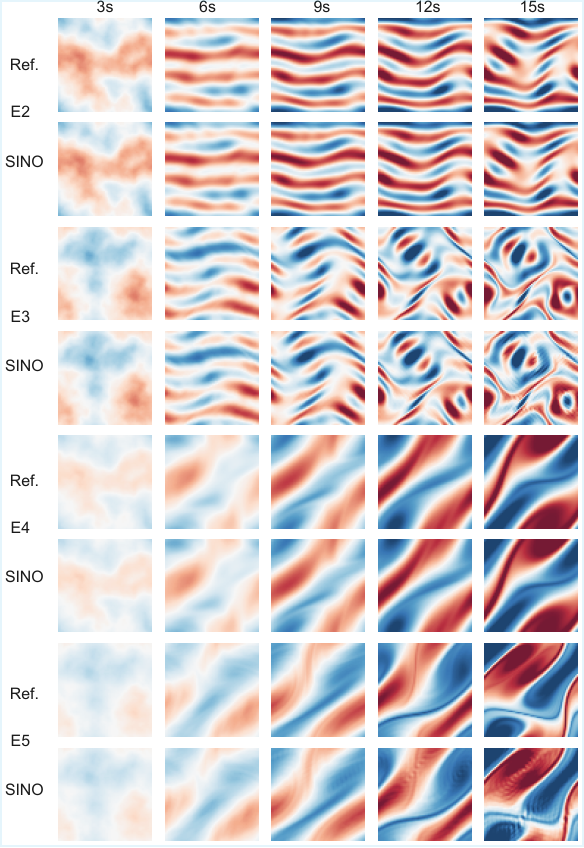}
    \caption{\textbf{Predicted trajectory of SINO.} Comparison of the vorticity field prediction trajectories for different scenarios (E2 to E5). The columns represent different prediction time steps at 3~s, 6~s, 9~s, 12~s, and 15~s.}
    \label{fig:snapshots}
\end{figure*}

In this section, we provide additional rollout trajectories to complement the additional results in the main text. Figs.~\ref{fig:IC0tra}-\ref{fig:IC3tra} further examine generalization across different initial conditions, including both in-distribution (IC0) and OOD (IC1-IC3) scenarios. Fig.~\ref{fig:snapshots} illustrates long-term predictions on the NSE cases (E2-E5).

\subsection{Annular Heat-Conduction Benchmark}
\label{app:annular_heat}

\begin{figure}
    \centering
    \includegraphics[width=0.99\linewidth]{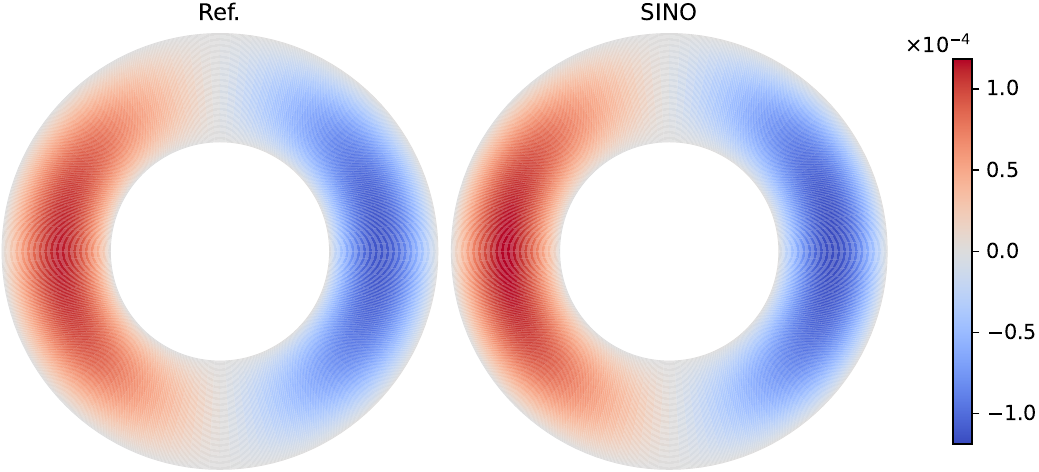}
    \caption{\textbf{Annular heat-conduction benchmark.}
Reference field and SINO prediction at a representative time step.}
    \label{fig:annular_heat}
\end{figure}

In this section, we provide an additional visualization of the annular heat-conduction benchmark to complement the quantitative results in the main text. This problem is defined on a non-Cartesian polar grid $(r,\theta)$ with radial Dirichlet boundary conditions, introducing both geometric complexity and non-periodic boundary constraints. Fig.~\ref{fig:annular_heat} shows the reference field and the corresponding SINO prediction at a representative time step on this benchmark. The visualization illustrates that SINO closely matches the reference solution.


\begin{figure*}
    \centering
    \includegraphics[width=1\linewidth]{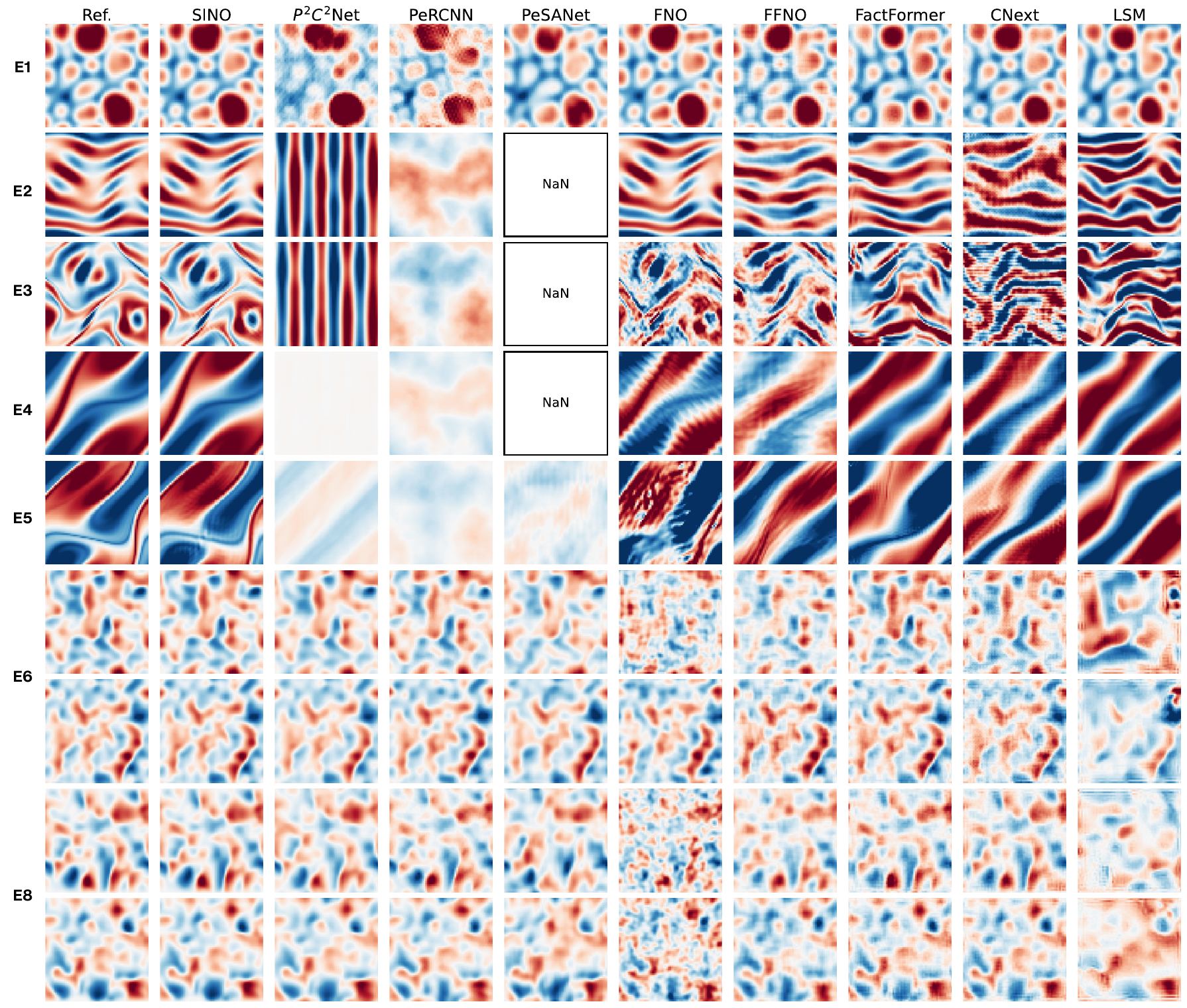}
    \caption{Full snapshots for E1-E8 (additional results beyond the main paper).}
    \label{fig:full}
\end{figure*}

\stopcontents[appendix]
\end{document}